\documentclass[11pt,a4paper]{article}

\usepackage[utf8]{inputenc}
\usepackage[T1]{fontenc}
\usepackage{charter}                 
\usepackage[scaled=0.92]{helvet}     
\usepackage{lmodern}                 

\usepackage[a4paper,top=1.05in,bottom=1.15in,left=1.35in,right=1.35in,%
            headheight=16pt,headsep=18pt,footskip=28pt]{geometry}

\usepackage{setspace}
\usepackage[skip=6pt plus1pt,indent=0pt]{parskip}  

\usepackage{amsmath,amssymb}
\usepackage{booktabs}
\usepackage{longtable}
\usepackage{array}
\usepackage{tabularx}
\usepackage{graphicx}
\usepackage{float}                   
\usepackage{placeins}                
\usepackage{tikz}
\usetikzlibrary{shapes.geometric,arrows.meta,positioning,fit,backgrounds}
\usepackage{xcolor}
\usepackage{tcolorbox}
\tcbuselibrary{skins,breakable}

\usepackage[font=small,labelfont={sf,bf},format=plain,justification=justified,%
            labelsep=period,skip=6pt]{caption}

\usepackage[round,authoryear]{natbib}

\usepackage[activate={true,nocompatibility},final,%
            tracking=true,kerning=true,spacing=true,factor=1100,stretch=10,shrink=10]{microtype}

\usepackage{fancyhdr}
\renewcommand{\headrulewidth}{0.3pt}
\renewcommand{\headrule}{\hbox to\headwidth{\color{black!35}\leaders\hrule height \headrulewidth\hfill}}

\fancypagestyle{plain}{%
  \fancyhf{}%
  \fancyfoot[C]{\footnotesize\color{black!55}\thepage}%
  \renewcommand{\headrulewidth}{0pt}%
}

\newcommand{\pkg}{\pkgname{pelloran-safety-lens}}  
\newcommand{\pkgcommit}{\texttt{9382839}}  
\newcommand{\pkgdoi}{\texttt{10.5281/zenodo.24345474}}  
\definecolor{pelBlue}{HTML}{4C78A8}
\definecolor{pelOrange}{HTML}{F58518}
\definecolor{pelLight}{HTML}{F0F4F8}
\definecolor{pelBorder}{HTML}{2C3E50}

\tikzset{
  stepbox/.style = {
    rectangle, rounded corners=3pt,
    draw=pelBorder, line width=0.6pt,
    fill=pelLight,
    text width=8.2cm, align=left,
    inner sep=6pt,
    font=\small
  },
  stepnum/.style = {
    circle, draw=pelBorder, line width=0.6pt,
    fill=pelBlue, text=white,
    minimum size=6mm, inner sep=0pt,
    font=\small\bfseries
  },
  flowarrow/.style = {
    -{Latex[length=2mm,width=2mm]},
    line width=0.6pt, draw=pelBorder
  },
  cellcol/.style = {
    rectangle, rounded corners=2pt,
    draw=pelBorder!30, line width=0.4pt,
    fill=pelLight!40,
    inner sep=4pt
  },
  cellhead/.style = {
    rectangle, rounded corners=2pt,
    draw=pelBorder, line width=0.6pt,
    fill=pelBlue, text=white,
    text width=3.1cm, align=center,
    inner sep=4pt,
    font=\small\bfseries\sffamily
  },
  cellstep/.style = {
    rectangle, rounded corners=2pt,
    draw=pelBorder!70, line width=0.4pt,
    fill=white,
    text width=3.1cm, align=center,
    inner sep=3pt,
    font=\scriptsize
  },
  proprietary/.style = {
    rectangle, rounded corners=2pt,
    draw=pelOrange, line width=0.7pt, dashed,
    fill=pelOrange!8,
    text width=3.1cm, align=center,
    inner sep=3pt,
    font=\scriptsize\itshape
  },
  outputbox/.style = {
    rectangle, rounded corners=2pt,
    draw=pelBorder, line width=0.5pt,
    fill=pelLight,
    text width=3.1cm, align=center,
    inner sep=3pt,
    font=\scriptsize\bfseries
  },
  cellarrow/.style = {
    -{Latex[length=1.5mm,width=1.5mm]},
    line width=0.5pt, draw=pelBorder!70
  }
}

\makeatletter
\renewcommand{\@seccntformat}[1]{\color{pelBlue}\csname the#1\endcsname\hspace{0.7em}}
\makeatother

\newtcolorbox{keyfindingbox}[1][]{%
  enhanced,
  colback=pelLight,
  colframe=pelBlue,
  boxrule=0pt,
  leftrule=3pt,
  arc=1pt, outer arc=1pt,
  fonttitle=\bfseries\sffamily,
  coltitle=pelBorder,
  title={Key finding},
  attach title to upper,
  after title={\ \ ---\ \ },
  left=8pt, right=8pt, top=6pt, bottom=6pt,
  #1
}
\newtcolorbox{contributionsbox}[1][]{%
  enhanced,
  colback=pelLight,
  colframe=pelOrange,
  boxrule=0pt,
  leftrule=3pt,
  arc=1pt, outer arc=1pt,
  fonttitle=\bfseries\sffamily,
  coltitle=white,
  title={Summary of contributions},
  left=8pt, right=8pt, top=6pt, bottom=6pt,
  #1
}
\definecolor{pelWarn}{HTML}{C77400}
\definecolor{pelWarnBg}{HTML}{FDF6EC}
\newtcolorbox{limitationsbox}[1][]{%
  enhanced, breakable,
  colback=pelWarnBg,
  colframe=pelWarn,
  boxrule=0pt,
  leftrule=3pt,
  arc=1pt, outer arc=1pt,
  fonttitle=\bfseries\sffamily,
  coltitle=white,
  title={Limitations at a glance},
  left=8pt, right=8pt, top=6pt, bottom=6pt,
  #1
}

\usepackage[colorlinks=true,linkcolor=pelBorder,urlcolor=pelBlue,%
            citecolor=pelBlue,pdftitle={Small Learned Projections for Semantic-Reversal-Aware Retrieval with Spectral Acceleration},%
            pdfauthor={Sayed Amir Karim},pdfborder={0 0 0}]{hyperref}

\newcolumntype{C}{>{\centering\arraybackslash}p{1.6cm}}
\newcommand{\HN}{\text{HN}@10}
\newcommand{\HitC}{\text{HitC}@10}
\newcommand{\pkgname}[1]{\textsf{\small #1}}

\makeatletter
\renewcommand{\maketitle}{%
  \begingroup
    \thispagestyle{plain}%
    \null\vspace{-0.6in}%
    \begin{center}%
      {\huge\bfseries\@title\par}%
      \vspace{14pt}%
      {\large\@author\par}%
      \vspace{16pt}%
      {\footnotesize\itshape\color{black!55}\@date\par}%
    \end{center}%
    \vspace{18pt}%
  \endgroup
}
\makeatother

\title{Small Learned Projections for Semantic-Reversal-Aware Retrieval\\with Spectral Acceleration\\[6pt]
       {\normalsize\mdseries\itshape A Four-Domain Empirical Study}}

\author{%
  Sayed Amir Karim\\[4pt]
  {\normalsize Pelloran Research}\\[1pt]
  {\normalsize\texttt{opensource@pelloran.com}}%
}

\date{v0.16 preprint \textperiodcentered\ 2026-11-13}

\begin{document}
\maketitle

\begin{abstract}
Dense-embedding retrievers can fail adversarially: when a query is
answered by document $A$ and reversed by a near-copy $B$ that differs from
$A$ by a small meaning-changing edit (a ``trap''), raw cosine over
E5-Large-v2 places $B$ in the top-10 for 82--100\,\% of queries at a pool
of $5\,000$. We study two things. First, a small learned projection
$W\colon \mathbb{R}^{1024} \to \mathbb{R}^{d}$ ($d \in \{11, 32\}$),
trained by a deterministic recipe on a declared set of cases, that
suppresses the trap: for the declared cases, $\HN = 0.0000$ (no trap in
the top-10) in all twelve $(\text{domain}, \text{pool-size})$ settings
across four independent domains---legal, medical, patent, financial---and
pools to $100\,000$, while the correct paraphrase is retained for
93--99\,\% of cases ($\HitC$), under exact nearest-neighbour search.
Second, a boundary-conditioned accelerator that serves the projection at
that scale: an acceleration of spectral graph retrieval
\citep{sahinoglu2026spectrag} that replaces a dense eigendecomposition
with a boundary-value solve, holding $\HN = 0.0000$ at $n = 100\,000$ in
every domain, retrieving the ${\sim}300$-query suite in $0.6$--$1.1$\,s
(${\sim}2$--$4$\,ms per query on average). Because the lens's output \emph{is}
the stored representation --- a few floats per document --- a hardened,
specialised index per domain or tenant is ${\sim}90\times$ smaller and
dimension-proportionally faster than the full embedding, making many small
declared indexes a practical alternative to one large general-purpose vector
store (\S\ref{sec:deployment}).

The claim is scoped to the declared cases on the measured pools, with no
probabilistic bound; the high $\HitC$ ($0.93$--$0.99$) is for those declared
cases, and drops for a reworded query against an unchanged corpus
($\HitC \approx 0.79$, trap suppression still holding). A projection transfers
only partly to other domains, which we report in full. The recipe, lenses, trap sets and evaluation
harness are released under AGPL-3.0; the accelerator's implementation is a
reserved commercial component, and every $\HN$ number is reproducible
under exact search.
\end{abstract}

\section{Introduction}

Adversarial trap corpora are the strongest available stress-test for a
dense-embedding retriever. Unlike random-negative distractors---which are
trivially rejected---a trap document $B$ differs from the correct answer
$A$ by a small meaning-changing edit---often a single token (\texttt{must}
vs.\ \texttt{must not}, \texttt{500\,mg} vs.\ \texttt{5000\,mg},
\texttt{comprising} vs.\ \texttt{consisting of}, \texttt{increased} vs.\
\texttt{decreased}), and a short multi-token edit in the legal corpus. Cosine
similarity in the embedding space assigns near-identical scores to $A$
and $B$, so $B$ routinely appears in the top-$k$ retrieved list for any
query anchored on $A$. In domains with regulatory or compliance stakes
(legal contracts, medical treatment guidelines, patent claim scope,
financial disclosures), a trap surfacing in top-10 is a silent
operational failure---the retriever confidently returns a document that
reverses the intended semantics of the query.

The dense retriever we measure (raw cosine over E5-Large-v2, no task
prefixes) is trap-vulnerable in this configuration: without a lens, the
trap reaches the top-10 for 82--100\,\% of queries at a pool of
$5\,000$ documents (\S\ref{sec:results}). Whether other encoders and
configurations behave the same is not measured here
(\S\ref{sec:limits}). Our contribution is a
training procedure that produces a small linear projection which keeps
the trap out of the top-10 for the cases it is declared for, without
collapsing retrieval utility (\S\ref{sec:declared}).

The mechanism is not novel in isolation. Learned linear projections
trained with a preservation loss plus a hard-negative hinge term are
standard building blocks in the metric-learning literature. What we
contribute is:

\begin{enumerate}
  \item \textbf{A precise construction recipe} that gives zero trap hits for the declared cases in every domain we
    measured, with the same code path.
  \item \textbf{Four-domain empirical validation}---legal, medical,
    patent, financial---with matched methodology and zero per-domain
    code changes.
  \item \textbf{A cross-domain transfer experiment.} Every trained
    lens is applied to the other domains' trap corpora. A lens is built
    for its own domain's cases; the experiment marks where it applies.
    Poor transfer is the expected outcome, but good transfer would not
    refute the recipe: a lens could also learn structure the domains
    share.
  \item \textbf{A complete $4\times 4$ transfer matrix} measuring
    twelve mismatched-lens/trap-corpus pairs. Eleven stay at
    $\HN \in [0.44, 0.85]$; one sits at $\HN = 0.2957$, below the
    $0.30$ the pre-registration had set, and is reported as missed.
  \item \textbf{Reproducibility infrastructure.} SHA-256-anchored
    artefacts, deterministic trap-generation scripts, a preflight
    duplicate-check guardrail, and an OSS release under AGPL-3.0.
\end{enumerate}

We report enough measurement detail that any reader with a modest
CPU-only machine and 24\,GB of RAM can reproduce every number in the
paper within a working day.

\begin{contributionsbox}
6-step recipe (\S\ref{sec:method}) $\to$ $\HN = 0.0000$ on four domains;
complete $4 \times 4$ transfer matrix (\S\ref{sec:matrix}): 11 of 12
off-diagonal cells at $\HN \in [0.44, 0.85]$; the 12\textsuperscript{th}
(Legal $d\!=\!11$ on Patent, $\HN = 0.2957$) is below the pre-registered
$0.30$ and reported as missed; a $d = 32$ retrain
(\S\ref{sec:ablation}) explains it but does not pass it. The trap sets, generators, lenses, trainer
and harness are released as \pkg{} under AGPL-3.0.
\end{contributionsbox}

\section{Related Work}

\paragraph{Dense retrieval baselines.} E5-Large-v2
\citep{wang2023e5}, BGE-Large \citep{chen2024bge}, and GTE-Large
\citep{li2023gte} are representative of the 1024-dimensional
sentence-encoder class, producing $L_2$-normalisable dense embeddings
with cosine as the natural similarity. In this paper we measure only
E5-Large-v2 on the four trap corpora; the underlying embedding
geometry that produces the trap vulnerability is shared across the
class (both A and B lie inside the same near-token-copy neighbourhood
in cosine space), so the same behaviour on BGE / GTE is a hypothesis
for further experiments; we have not measured it (see Limitations,
\S\ref{sec:limits}).

\paragraph{Hybrid dense--sparse retrieval.} BM25/dense hybrids
\citep{robertson2009bm25,karpukhin2020dpr} and ColBERT-style
late-interaction models \citep{khattab2020colbert} are not, in
general, a defence against the specific trap class studied here: our
$B$ differs from $A$ by a single semantically-reversing token
(\texttt{must} vs.\ \texttt{must not}), so BM25 assigns $A$ and $B$
near-identical lexical scores and hybrid dense--sparse retrievers
inherit the dense component's trap-fall. We report Cell A (raw cosine)
as the dense-only baseline and do not run a BM25 experiment because
we do not expect it to move on this trap taxonomy; a hybrid comparison
on trap classes with wider lexical spread is scoped for follow-up
work, not claimed here.

\paragraph{Adversarial IR.} The adversarial-retrieval literature is
dominated by black-box attack generation on the offensive side
\citep{guo2019blackbox,song2020adversarial,wu2022prada}. Our work sits
on the defense side: given a known trap taxonomy, how to train a
projection that suppresses trap-fall while preserving utility.

\paragraph{Negation in neural retrieval (NevIR).} \citet{weller2024nevir}
build a benchmark of document pairs that differ only by a negation and
ask retrieval models to rank each pair for a query. Most models rank at
or below chance; cross-encoders do best, then late-interaction models,
with bi-encoders and sparse models last; fine-tuning on contrastive
negation data helps but leaves a large gap. Our trap pairs are of the
same kind---one meaning-reversing difference---but cover more contrasts
than negation (quantities, claim scope, direction), and our remedy is a
small projection on top of a frozen bi-encoder rather than a different
retriever architecture. We do not evaluate on NevIR here
(\S\ref{sec:limits}).

\paragraph{Metric learning and steering vectors.} Preservation-plus-hinge
losses are standard in metric learning
\citep{xing2003metric,schroff2015facenet,sohn2016npair}. Steering vectors
for language models \citep{turner2023activation,zou2023repe} apply
similar structural interventions on model activations. Our lens is
closest to a supervised linear steering projection trained on a small
$A/B/C$ triple set with the hinge term operating in the projected space.

\paragraph{Spectral graph retrieval (SpectRAG).}
\citet{sahinoglu2026spectrag} propose SpectRAG, a spectral
graph-augmented retriever for multi-hop retrieval-augmented generation.
It builds a weighted graph over document chunks whose edges encode
semantic similarity, document adjacency, section hierarchy, shared
entities, citation links, and similarity between LLM-generated utility
questions, and it selects the final context under a fixed token budget.
We build directly on this formulation. The serving layer of our
pipeline (\S\ref{sec:accel}) is an acceleration of SpectRAG---the same
spectral-retrieval formulation, with the dense eigendecomposition
replaced by a boundary-conditioned equilibrium solve---which lifts it
from a few thousand chunks to the $10^5$-document pools used here. We do
not re-run the original dense SpectRAG at these scales (it does not reach
them); we cite it as the method we extend.

\paragraph{Duplicate contamination as a rigour gate.} Benchmark
inflation from duplicate near-copies is a well-documented failure mode
in NLP evaluation \citep{dodge2021c4,lee2022dedup}. Our data-quality
audit (\S\ref{sec:preflight}) surfaces a 22\,\% duplicate rate in an
early version of our legal trap corpus that inflated $\HitC$ by
16\,pp. We ship the resulting preflight guardrail as a hard gate
before any lens training run.

\section{Method}
\label{sec:method}

\subsection{What a lens is declared for}
\label{sec:declared}

A \emph{case} is one specific contrasting statement pair: a statement
$A$, its trap $B$ ($A$ with one meaning-reversing change), and a
paraphrase $C$ of $A$. For example, $A$ ``Metformin is dosed at
500\,mg twice daily for adults with type-2 diabetes'', $B$ the same
sentence with 5000\,mg, and $C$ ``Adults with type-2 diabetes receive
metformin 500\,mg BID''. A case is \emph{not} a contrast type: ``dosage
flips'' in general is a type; the metformin pair is a case.

A lens is declared for a set of cases---an insurer's contract terms
AA, BB and CC, say---and is trained on exactly those. For a declared
case it is meant to keep $B$ out of the top-10 for a query anchored on
$A$, and $C$ in it, in the evaluated distractor pools. Cases
outside the declared set---including new statement pairs of the same
type---are not covered by design: one declares and trains a lens for
them (DD, EE), combines lenses, or declares the whole set (AA to EE)
from the start.

A trap is unwanted where the reader needs the statement as it stands:
answering, compliance and clause retrieval, where returning ``must
not'' for ``must'' reverses the answer. Where the contradiction itself
is sought---a search for conflicting clauses or for a change between
versions---$B$ is a relevant result, and no lens is applied.

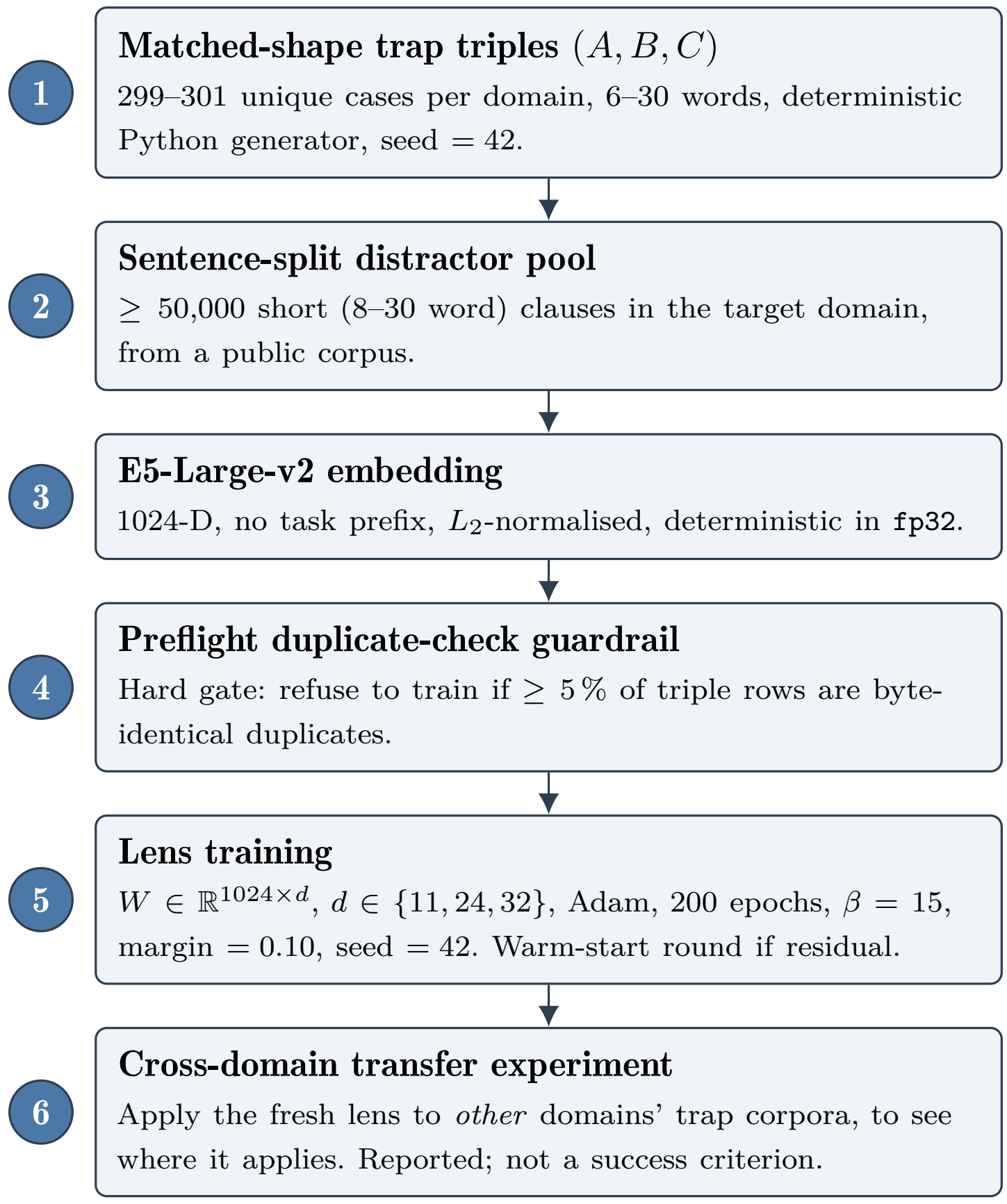
\begin{figure}[h]
\centering
\begin{tikzpicture}[node distance=4mm]
  \node[stepbox] (s1) {%
    \textbf{Matched-shape trap triples $(A, B, C)$} \\[1pt]
    {\scriptsize 299--301 unique cases per domain, 6--30 words, deterministic Python generator, seed $= 42$.}};
  \node[stepnum, left=2mm of s1.west] {1};
  \node[stepbox, below=of s1] (s2) {%
    \textbf{Sentence-split distractor pool} \\[1pt]
    {\scriptsize $\geq 50{,}000$ short (8--30 word) clauses in the target domain, from a public corpus.}};
  \node[stepnum, left=2mm of s2.west] {2};
  \node[stepbox, below=of s2] (s3) {%
    \textbf{E5-Large-v2 embedding} \\[1pt]
    {\scriptsize 1024-D, no task prefix, $L_{2}$-normalised, deterministic in \texttt{fp32}.}};
  \node[stepnum, left=2mm of s3.west] {3};
  \node[stepbox, below=of s3] (s4) {%
    \textbf{Preflight duplicate-check guardrail} \\[1pt]
    {\scriptsize Hard gate: refuse to train if $\geq 5\,\%$ of triple rows are byte-identical duplicates.}};
  \node[stepnum, left=2mm of s4.west] {4};
  \node[stepbox, below=of s4] (s5) {%
    \textbf{Lens training} \\[1pt]
    {\scriptsize $W \in \mathbb{R}^{1024 \times d}$, $d \in \{11, 24, 32\}$, Adam, 200 epochs, $\beta = 15$, margin $= 0.10$, seed $= 42$. Warm-start round if residual.}};
  \node[stepnum, left=2mm of s5.west] {5};
  \node[stepbox, below=of s5] (s6) {%
    \textbf{Cross-domain transfer experiment} \\[1pt]
    {\scriptsize Apply the fresh lens to \emph{other} domains' trap corpora, to see where it applies. Reported; not a success criterion.}};
  \node[stepnum, left=2mm of s6.west] {6};
  \draw[flowarrow] (s1) -- (s2);
  \draw[flowarrow] (s2) -- (s3);
  \draw[flowarrow] (s3) -- (s4);
  \draw[flowarrow] (s4) -- (s5);
  \draw[flowarrow] (s5) -- (s6);
\end{tikzpicture}
\caption{The six-step lens training recipe. Steps 1--3 prepare the
domain data; step 4 is the rigour guardrail; step 5 trains the safety
lens; step 6 is the transfer experiment.}
\label{fig:pipeline}
\end{figure}

Figure~\ref{fig:pipeline} summarises the six deterministic steps of the
recipe. Every step is described in detail in the subsections that follow.

\subsection{Trap corpus construction}

Every trap corpus consists of $N$ unique test cases, each a triple
$(A, B, C)$:

\begin{itemize}
  \item $A$ --- the anchor. A factually correct short clause (6--30
    words) in the target domain's vocabulary.
  \item $B$ --- the trap. Identical to $A$ except for one
    meaning-flipping token (or a small closed set of tokens for
    compound trap types). Length within $\pm 20\,\%$ of $A$.
  \item $C$ --- the utility target. A paraphrase of $A$ preserving
    intent, using different surface phrasing.
\end{itemize}

\paragraph{Shape invariants.} All three of $A, B, C$ are within
6--30 words, with the max-to-min word-count ratio bounded by $2.0$. In the
medical, patent and financial corpora $B$ differs from $A$ by one
meaning-flipping token; in the broader legal corpus the change is a short
meaning-altering edit (one to a few tokens). $C$ uses different phrasing
but preserves the anchor concept.

\paragraph{Domain-specific taxonomies.} The medical, patent and financial
corpora each use four approximately-balanced single-token flip types; the
legal corpus uses a broader, unbalanced taxonomy of 23 contract-clause
trap types (the five largest shown below). The table groups flips by
(type, example edit):

\begin{center}
\small
\begin{tabular}{@{}l@{\hspace{1em}}l@{\hspace{0.8em}}l@{}}
\toprule
\textbf{Domain} & \textbf{Flip type} & \textbf{Example pair} \\
\midrule
Legal      & numeric unit (102)        & a measurement-unit edit \\
           & numeric threshold (46)    & a threshold / boundary edit \\
           & temporal scope (45)       & a date / window edit \\
           & anchor format (33)        & a clause-reference edit \\
           & conditional trigger (33)  & \texttt{must}\,$\leftrightarrow$\,\texttt{must not} \\
\addlinespace[2pt]
Medical    & dosage                    & \texttt{500\,mg}\,$\leftrightarrow$\,\texttt{5000\,mg} \\
           & contraindication          & \texttt{indicated}\,$\leftrightarrow$\,\texttt{contraindicated} \\
           & population                & \texttt{adults}\,$\leftrightarrow$\,\texttt{pediatric} \\
           & drug substitution (LASA)  & \texttt{metformin}\,$\leftrightarrow$\,\texttt{metoprolol} \\
\addlinespace[2pt]
Patent     & claim-scope               & \texttt{comprising}\,$\leftrightarrow$\,\texttt{consisting of} \\
           & enablement                & \texttt{enabled}\,$\leftrightarrow$\,\texttt{disabled} \\
           & art                       & \texttt{novel}\,$\leftrightarrow$\,\texttt{prior art} \\
           & mechanism substitution    & \texttt{encrypt}\,$\leftrightarrow$\,\texttt{decrypt} \\
\addlinespace[2pt]
Financial  & direction                 & \texttt{increased}\,$\leftrightarrow$\,\texttt{decreased} \\
           & polarity                  & \texttt{gain}\,$\leftrightarrow$\,\texttt{loss} \\
           & polarity                  & \texttt{asset}\,$\leftrightarrow$\,\texttt{liability} \\
           & period                    & \texttt{Q1}\,$\leftrightarrow$\,\texttt{Q4} \\
           & party                     & \texttt{lender}\,$\leftrightarrow$\,\texttt{borrower} \\
\bottomrule
\end{tabular}
\end{center}

Each corpus contains ${\sim}300$ unique triples---${\sim}75$ per type for
the medical, patent and financial domains; the legal set spreads its 299
cases across 23 types (largest 102, several with only a handful). We
construct them via deterministic Python generators, shipped in the OSS
release as \texttt{scripts/gen\_*\_triples\_*.py}.

\subsection{Distractor pool}

For each domain we assemble a distractor pool of $\geq 50\,000$ short
(8--30 word) clauses in the target domain, sentence-split from a
public corpus. \textbf{Matched-shape distractors are essential}
---mixing sentence-length $A/B/C$ with paragraph-length distractors
makes the retrieval geometry noise-dominated (Appendix~\ref{app:failures},
failure mode~3).

Sources per domain:

\begin{itemize}
  \item \textbf{Legal}: LexGLUE, sentence-split.
  \item \textbf{Medical}: PubMed abstracts, sentence-split,
    ${\sim}200\,000$ unique.
  \item \textbf{Patent}: HuggingFace BIGPATENT abstracts, sentence-split
    across eight USPTO categories, 182\,335 unique.
  \item \textbf{Financial}: five HuggingFace financial NLP datasets
    combined and sentence-split, 177\,561 unique.
\end{itemize}

\subsection{Embedding}

We use \texttt{intfloat/e5-large-v2} \citep{wang2023e5}. No task prefix
(\texttt{passage:} / \texttt{query:}) applied. $L_2$-normalisation after
encoding. 1024-dimensional output. The training procedure is
architecture-agnostic; the choice of E5 here is representative rather
than load-bearing.

\subsection{Preflight duplicate-check guardrail}
\label{sec:preflight}

Before training, we compute SHA-256 hashes of every triple's flattened
\texttt{f32} embedding bytes and refuse to proceed if $\geq 5\,\%$ of
case rows are byte-identical duplicates. This is a hard gate. In an
early experiment on the legal corpus we discovered a 22\,\% duplicate
rate that had silently inflated $\HitC$ by 16 percentage points; the
guardrail enforces the discipline at generation time. Preflight-check
code and threshold rationale ship in the OSS release.

\subsection{Lens training}

The lens is a matrix $W \in \mathbb{R}^{1024 \times d}$ with
$d \in \{11, 24, 32\}$. Training minimises:
\[
  \mathcal{L}(W) = \alpha \cdot \mathcal{L}_{\text{preserve}}(W)
                 + \beta  \cdot \mathcal{L}_{\text{hinge}}(W)
                 + \lambda \cdot \operatorname{mean}(W^{2})
\]
where:

\begin{itemize}
  \item $\mathcal{L}_{\text{preserve}}(W)$ is the mean-squared error
    between the pre- and post-projection pairwise cosine similarities,
    over a sampled batch of \emph{distinct} document pairs (all $i < j$,
    not only positive pairs). It preserves the geometry that makes
    retrieval work.
  \item $\mathcal{L}_{\text{hinge}}(W)$ is the per-triple
    $\max\bigl(0,\; \cos(A_{\text{proj}}, B_{\text{proj}})
                    - \cos(A_{\text{proj}}, C_{\text{proj}})
                    + \text{margin}\bigr)$. It pushes the trap $B$
    away from the query $A$ in the projected space while keeping the
    utility target $C$ close.
  \item $\alpha = 1.0$, $\beta \in \{10, 15, 17\}$,
    $\text{margin} \in \{0.05, 0.10, 0.12\}$,
    $\lambda = 1\!\times\!10^{-4}$. The regulariser is the \emph{mean} of
    the squared entries of $W$, not a summed norm, so $\lambda$ is scaled
    to that convention.
\end{itemize}

Optimiser: Adam, learning rate $5\!\times\!10^{-3}$,
$\text{pairs\_per\_epoch} = 8000$, $\text{epochs} = 200$, $\text{seed} = 42$
(fully deterministic). Warm-starting from a prior lens
(\texttt{--init-aqwt}) shortens convergence and closes the last
percentage points of \texttt{hn\_loss}. For a new domain from
cold-start, $d = 32$ and $\beta = 15$ typically produces near-zero
\texttt{hn\_loss}; a second warm-start round with the same configuration
or $\beta = 17$ eliminates the residual. $\beta = 20$ destabilises
training (see Appendix~\ref{app:hyperparam}).

Training wall-clock: approximately 90\,s per lens on a modern multi-core
CPU (an AMD EPYC server, no GPU). No gradient checkpointing or mixed
precision required.

\subsection{Cross-domain transfer experiment}

The final step of the recipe. Take the freshly trained domain-$X$ lens
and measure it on the trap corpora of the other domains. A lens is built
for its own domain's cases, so we expect it to do much worse there; the
experiment shows where a lens applies. It is not a validity test: a lens
that also separated another domain's traps would have learned structure
the domains share, not failed. Our pre-registration had set
$\HN \geq 0.30$ off-domain as a criterion. We report that criterion as
it was written (\S\ref{sec:matrix}), but no longer treat it as a
condition for the recipe's validity.

\subsection{Determinism and reproducibility infrastructure}

Every step of the recipe is deterministic given a fixed seed and pinned
dependencies:

\begin{itemize}
  \item Trap generators emit the identical 301-triple corpus for a
    given $\text{seed} = 42$.
  \item E5 embedding is deterministic in \texttt{fp32}.
  \item Preflight duplicate check is a pure hash operation.
  \item Adam with fixed seed is deterministic in CPU-only mode.
  \item The measurement harness reports raw counts, not probabilistic
    estimates.
\end{itemize}

Every artefact (trap corpus, distractor pool, lens weight matrix,
measurement JSON) is anchored by SHA-256 in
Appendix~\ref{app:sha-anchors}. Re-running the pipeline with the same
seed and pinned dependencies reproduces the SHA exactly.

\subsection{Serving at scale: a boundary-conditioned spectral retrieval accelerator}
\label{sec:accel}

The lens is a reranking primitive. To serve it over pools of
$10^5$ documents we pair it with a spectral retrieval accelerator.
SpectRAG \citep{sahinoglu2026spectrag} selects context by a spectral
analysis of a similarity graph over document chunks. Building that
spectral structure densely requires an eigendecomposition of the graph
Laplacian---an \emph{offline}, per-corpus cost of $O(n^3)$ that, in our
measurements, exhausts memory well before $n = 100\,000$ and so caps the
corpus size a dense build can index.

Our accelerator keeps the spectral-retrieval formulation but avoids the
dense eigendecomposition. Writing the graph Laplacian in block form with a
small boundary set $B$ (the query-anchored seeds) and interior $I$, the
equilibrium field over the interior is the discrete harmonic extension
\[
   T_I^{\mathrm{eq}} = -\,L_{II}^{-1}\, L_{IB}\, T_B
   \qquad\Longleftrightarrow\qquad
   L_{II}\, x = -\,L_{IB}\, T_B ,
\]
the Dirichlet solution (Schur complement) of the boundary conditions. The
interior field $T_I^{\mathrm{eq}}$ scores the interior nodes and the
retrieval ranking is read off from it, so one boundary-value solve
replaces the per-query spectral scoring.

\paragraph{Fixed boundary, varying values.} The boundary set is fixed at
the corpus level (a small set of anchor vertices); a query sets only the
boundary \emph{values} $T_B$---its similarities to the boundary
documents---changing the right-hand side $-L_{IB}\,T_B$ but not $L_{II}$.
The factorisation of $L_{II}$ is therefore computed once per corpus and
reused by every query, which is a solve against that fixed factorisation.

Three costs separate:

\begin{itemize}
  \item \textbf{Graph build} --- assembling the sparse Laplacian over the
    pool: offline, per corpus.
  \item \textbf{Factorisation} of $L_{II}$ --- offline, per corpus,
    amortised over all queries.
  \item \textbf{Per-query solve} --- the boundary setup and the triangular
    or iterative solves that produce $T_I^{\mathrm{eq}}$.
\end{itemize}

\paragraph{A conditional complexity claim.} We do not claim near-linearity
unconditionally. \emph{If} $L_{II}$ is banded with a bandwidth $b$ bounded
independently of $n$, a banded Cholesky factorisation and each triangular
solve are $O(n\,b^2)$---linear in $n$ for fixed $b$; \emph{equivalently},
an unfactorised Krylov solve costs $O(n\,s\,m)$ for sparsity $s$ and
iteration count $m$, near-linear when $s$ and $m$ stay bounded on the
retrieval graph (with preconditioning where needed). Our engine's
construction keeps the interior Laplacian in this bounded-bandwidth
regime; \emph{that} construction is the proprietary part and is not
disclosed. The claim is therefore conditional on the stated
bounded-bandwidth (resp.\ bounded-iteration) assumption---not a theorem
about arbitrary retrieval graphs; the measured wall-clock is in
\S\ref{sec:celld}.

The accelerator is an optional, swappable backend. The lens emits
$L_2$-normalised $d$-dimensional vectors; of the two backends we measure
---the accelerator and the exact-nearest-neighbour substitute shipped in
\pkg{}---both return the same $\HN$ on these pools. (A projection does not
by itself guarantee that an \emph{arbitrary} downstream retriever selects
the same documents; we claim this only for the two backends we compared.)
The accelerator supplies the latency and the $100\,000$-document reach
(Cell~D, \S\ref{sec:celld}); the safety property does not depend on it.

\section{Cell-based Measurement Framework}
\label{sec:cells}

We measure retrieval quality with four cells per pool-size configuration:

\begin{itemize}
  \item \textbf{Cell A} --- raw E5 cosine, no lens, no graph. The
    baseline every dense-retrieval system starts from.
  \item \textbf{Cell B} --- 3-stage cascade: a utility projection ($W_u$)
    $\to$ top-$k$ $\to$ lens $\to$ top-10. Pure projections, no graph.
  \item \textbf{Cell C} --- lens projection with \emph{exact
    nearest-neighbour} search over the projected pool. This is the open,
    reproducible safety measurement: every $\HN$ number in
    \S\ref{sec:results} is from this cell, and it reproduces in \pkg{}.
  \item \textbf{Cell D} --- the same lens \emph{served by the
    spectral accelerator} of \S\ref{sec:accel} (the deployed pipeline).
    $\HN$ is identical to Cell~C; Cell~D adds the $0.6$--$1.1$\,s suite
    retrieval (${\sim}2$--$4$\,ms per query) and the $100\,000$-document reach
    of \S\ref{sec:celld}. Its numbers are
    measured and SHA-anchored; its retriever is a reserved commercial
    component.
\end{itemize}

For each cell we report:

\begin{itemize}
  \item \HN{} --- hard-negative rate at $k=10$, defined as the
    fraction of queries where the trap $B$ appears in the top-10
    retrieved documents.
  \item \HitC{} --- correct-answer rate at $k=10$, defined as the
    fraction of queries where the paraphrase $C$ appears in the top-10.
\end{itemize}

Cell D is the deployed pipeline; Cell C is its open, exact-search
reproduction, with identical $\HN$. The domain-agnostic claim is carried
by Cells C and D together. Cells A and B are reference points.

\paragraph{Scope note.} The spectral accelerator underlying Cell D is
out-of-scope for this paper; only its interface is documented (the
mathematics of why it is near-linear is given in \S\ref{sec:accel}). The
lens produced by the recipe in \S\ref{sec:method} operates
independently of this retriever and can be paired with any retrieval
backend that consumes $L_2$-normalised $d$-dimensional embeddings.
Readers reproducing the paper use the exact-nearest-neighbour backend on
the projected pool (Cell~C): every $\HN$ number in \S\ref{sec:results} is
measured that way, with the package's exact search. The accelerator's
latency (Cell~D) requires the commercial retriever or an equivalent
near-linear spectral backend.

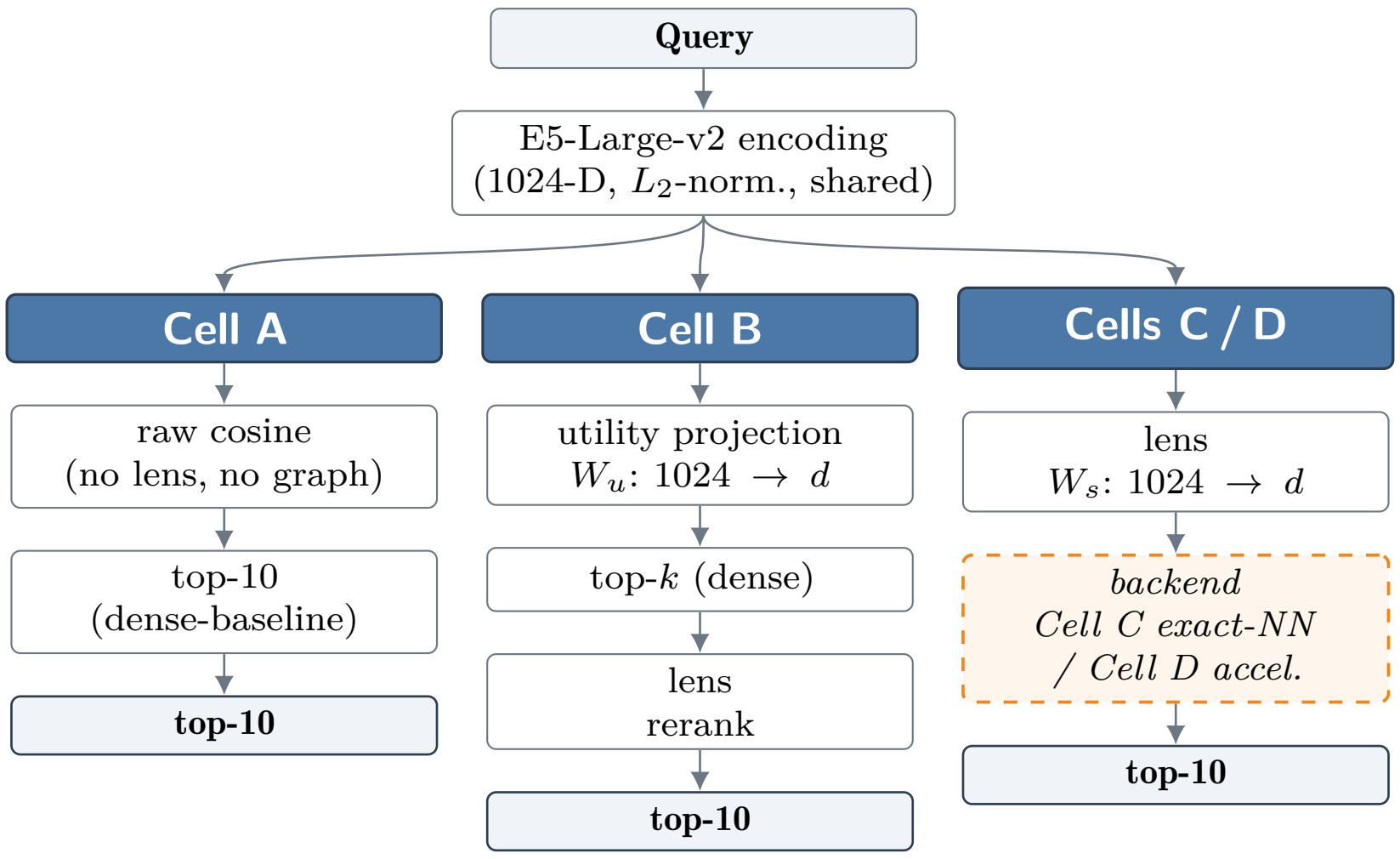
\begin{figure}[h]
\centering
\begin{tikzpicture}[node distance=3.2mm]
  \node[cellstep, fill=pelLight, font=\scriptsize\bfseries]
    (query) {Query};
  \node[cellstep, below=of query, text width=3.7cm]
    (embed) {E5-Large-v2 encoding\\{\scriptsize (1024-D, $L_2$-norm., shared)}};
  \draw[cellarrow] (query) -- (embed);

  \node[cellhead, below=6mm of embed, xshift=-37.3mm] (headA) {Cell A};
  \node[cellhead, right=3mm of headA] (headC) {Cell B};
  \node[cellhead, right=3mm of headC] (headD) {Cells C\,/\,D};

  \node[cellstep, below=of headA] (a1) {raw cosine\\{\scriptsize (no lens, no graph)}};
  \node[cellstep, below=of headC] (c1) {utility projection\\{\scriptsize $W_u$: $1024 \to d$}};
  \node[cellstep, below=of headD] (d1) {lens\\{\scriptsize $W_s$: $1024 \to d$}};

  \node[cellstep, below=of a1] (a2) {top-10\\{\scriptsize (dense-baseline)}};
  \node[cellstep, below=of c1] (c2) {top-$k$ (dense)};
  \node[proprietary, below=of d1] (d2) {backend\\{\scriptsize Cell~C exact-NN / Cell~D accel.}};

  \node[cellstep, below=of c2] (c3) {lens\\rerank};
  \node[outputbox, below=of a2] (aout) {top-10};
  \node[outputbox, below=of c3] (cout) {top-10};
  \node[outputbox, below=of d2] (dout) {top-10};

  \draw[cellarrow] (headA) -- (a1);
  \draw[cellarrow] (a1) -- (a2);
  \draw[cellarrow] (a2) -- (aout);

  \draw[cellarrow] (headC) -- (c1);
  \draw[cellarrow] (c1) -- (c2);
  \draw[cellarrow] (c2) -- (c3);
  \draw[cellarrow] (c3) -- (cout);

  \draw[cellarrow] (headD) -- (d1);
  \draw[cellarrow] (d1) -- (d2);
  \draw[cellarrow] (d2) -- (dout);

  \draw[cellarrow] (embed.south) .. controls +(0,-4mm) and +(0,4mm) .. (headA.north);
  \draw[cellarrow] (embed.south) .. controls +(0,-4mm) and +(0,4mm) .. (headC.north);
  \draw[cellarrow] (embed.south) .. controls +(0,-4mm) and +(0,4mm) .. (headD.north);
\end{tikzpicture}
\caption{The measurement cells. All share the same E5-Large-v2 embedding
of the query and differ in the operations applied to the pool before the
top-10 is returned. Cell A is the dense-retrieval baseline; Cell B is the
lens-only 3-stage cascade; the lens column feeds two interchangeable
backends---Cell~C (exact nearest-neighbour, open and reproducible) and
Cell~D (the spectral accelerator, reserved commercial; dashed box)---which
return the same $\HN$ (see §\ref{sec:cells} scope note).}
\label{fig:cells}
\end{figure}

Figure~\ref{fig:cells} contrasts the cells by the operations each
applies between the shared E5-Large-v2 embedding of the query and the
returned top-10 documents.

\section{Experiments}
\label{sec:experiments}

\subsection{Domains and sizes}

\begin{table}[h]
\centering
\footnotesize
\begin{tabular}{lr>{\raggedright\arraybackslash}p{3.1cm}cl}
\toprule
\textbf{Domain} & \textbf{Query cases} & \textbf{Distractor pool} & \textbf{Lens dim} & \textbf{Trap taxonomy} \\
\midrule
Legal                     & 299 & LexGLUE sentence-split                 & 11 & 23 types (largest 102) \\
Medical                   & 300 & PubMed sentence-split, ${\sim}200$k    & 32 & 4 $\times$ 75 \\
Patent (USPTO)            & 301 & BIGPATENT sentence-split, 182k         & 32 & 4 $\times$ 75 \\
Financial (SEC-adjacent)  & 301 & 5 HF financial datasets, 178k          & 32 & 4 $\times$ 75 \\
\bottomrule
\end{tabular}
\end{table}

Every corpus passes the $0.00\,\%$ preflight duplicate gate.

\subsection{Pool sizes}

Every domain is measured at approximately $n = 5\text{k}$,
$50\text{k}$, and $100\text{k}$ (query cases $+$ a same-domain
distractor slice); exact pool sizes vary by a few hundred rows
depending on how the underlying dedup-guardrailed AQTC binary is
sized per domain, and are reported explicitly in the $n$ column of
Table~1 in \S\ref{sec:results}. The pools are nested: the package's
pool builder (\texttt{embed\_and\_pack.py}) cuts each pool as the first
$N$ documents of one fixed distractor list, for Medical, Patent and
Financial; the Legal pools come from an earlier build whose record does
not state it.

\subsection{Hardware}

All Cell C measurements on a single M4 laptop (10 CPU cores, no GPU).
Lens training on an AMD EPYC server (CPU-only). No specialised hardware
required.

\section{Results}
\label{sec:results}

\subsection{The lens on its declared cases: four domains, three pool sizes (exact nearest-neighbour search)}

The headline table. Each row is one $(\text{domain}, \text{pool-size})$
pair, for the lens's declared cases (\S\ref{sec:declared}).
$\HN = 0.0000$ means that for none of the ${\sim}300$ declared cases
does the trap enter the top-10, among a pool of documents the lens never
saw; the declared cases' own phrasings are the trained ones. The result
is zero observed trap hits on the declared training cases across the
measured pools. We state no probabilistic bound on it: the cases are the
ones the lens was trained and selected on, not an independent sample.
The three pool sizes reuse the same cases.

\begin{table}[h]
\centering
\small
\caption{Cell~C --- the lens under exact nearest-neighbour search:
$\HN$ and $\HitC$ at $k = 10$ for the declared cases, four domains
$\times$ three pool sizes, with raw cosine ($\HitC$, Cell~A) for
reference. The open, reproducible measurement (\pkg{},
\texttt{results/pool\_sizes/}).}
\label{tab:results}
\begin{tabular}{lrccc}
\toprule
\textbf{Domain} & $n$ & $\HN$ (lens) & $\HitC$ (lens) & $\HitC$ (raw cosine) \\
\midrule
Legal     & 5\,897 & \textbf{0.0000} & 0.9866\rlap{\textsuperscript{$\dagger$}} & 0.6722 \\
Legal     & 50\,897 & \textbf{0.0000} & 0.9866\rlap{\textsuperscript{$\dagger$}} & 0.6722 \\
Legal     & 100\,897 & \textbf{0.0000} & 0.9833 & 0.6722 \\
Medical   & 5\,900 & \textbf{0.0000} & 0.9733 & 0.9867 \\
Medical   & 50\,900 & \textbf{0.0000} & 0.9700 & 0.9867 \\
Medical   & 100\,900 & \textbf{0.0000} & 0.9700 & 0.9867 \\
Patent    & 5\,903 & \textbf{0.0000} & 0.9701 & 0.9601 \\
Patent    & 50\,903 & \textbf{0.0000} & 0.9635 & 0.9601 \\
Patent    & 100\,903 & \textbf{0.0000} & 0.9568 & 0.9601 \\
Financial & 5\,903 & \textbf{0.0000} & 0.9601 & 1.0000 \\
Financial & 50\,903 & \textbf{0.0000} & 0.9468 & 1.0000 \\
Financial & 100\,903 & \textbf{0.0000} & 0.9269 & 1.0000 \\
\bottomrule
\end{tabular}
\end{table}

\noindent
Every number in the table is from the reproducibility package's own
measurement (\pkg{}, \texttt{results/pool\_sizes/},
commit \pkgcommit, Zenodo \pkgdoi; the SHA-256 of each result file is in
Appendix~\ref{app:sha-anchors}), with exact nearest-neighbour search
over the projected pool. The pool $n$ is the declared cases' own rows
plus the distractors: $n = 3 \times \text{cases} + \text{distractors}$
(the query's $A$, the trap $B$ and the paraphrase $C$ of every case).
This table is Cell~C (exact nearest-neighbour search). The served
pipeline (Cell~D, Table~\ref{tab:celld}) runs the accelerator and reports
the corresponding $\HitC$, lower by up to ${\sim}0.03$ at the largest pool
(Patent at $100\text{k}$: $0.9269$ under the accelerator vs.\ $0.9568$
here); $\HN = 0.0000$ in both. Cell~B (the lens-only cascade) is tabulated in
the same-pool runtime comparison (Table~\ref{tab:samepool},
\S\ref{sec:celld})---and is not part of \pkg{}. The last
column is Cell A's $\HitC$: raw cosine on the same pools, the same
encoder, no E5 prefixes, measured by the same exact search
(\texttt{miss\_diagnostic.py --raw}; \texttt{results/raw\_cosine/},
hashes in Appendix~\ref{app:sha-anchors}). Against raw cosine the lens
raises $\HitC$ in Legal at every size ($0.6722 \to 0.9866$ at
$5\text{k}$, $0.9833$ at $100\text{k}$). In Patent it raises it at
$5\text{k}$ ($0.9601 \to 0.9701$) and $50\text{k}$ ($0.9601 \to 0.9635$),
and lowers it at $100\text{k}$ ($0.9568$). It costs some in Medical
($0.9867 \to 0.9733$ at $5\text{k}$) and Financial ($1.0000 \to 0.9601$).

\textsuperscript{$\dagger$}\,The identical Legal $\HitC = 0.9866$ at
$n = 5\text{k}$ and $n = 50\text{k}$ is not a copy: the same four
paraphrase misses land outside top-10 at both pool sizes; the fifth
miss only appears at $n = 100\text{k}$ where the additional 50k
distractors dilute the ranking enough to push one borderline $C$
below rank 10. HitC on Legal is more robust to pool growth than on
the other three domains because the LexGLUE distractor pool is
lexically farther from the legal A/C pairs than
PubMed / BIGPATENT / financial-news are from their in-domain triples.
nDCG@10 is not reported here because our miss-diagnostic harness does
not compute the full ranking required and cross-pipeline nDCG numbers
from our earlier runs are inconsistent; readers wanting nDCG can
substitute an exact-NN backend plus a full-ranking metric library on
the projected pool.

\paragraph{What grows with the pool.} With a fixed projection, exact
nearest-neighbour search and unchanged document vectors, adding
documents to a nested pool can only keep a trap's rank or push it down.
A trap outside the top-10 at $5\text{k}$ therefore stays outside it at
$50\text{k}$ and $100\text{k}$, so for the three nested domains the
stable $\HN = 0.0000$ across the growing pools is expected, close to
trivial; for Legal it is measured at each size. The informative scaling
result is $\HitC$: the correct paraphrase has to survive twenty times
more competition. From $5\text{k}$ to $100\text{k}$ it is retained for
$98.66 \to 98.33\,\%$ of the cases in Legal, $97.33 \to 97.00\,\%$ in
Medical, $97.01 \to 95.68\,\%$ in Patent, and $96.01 \to 92.69\,\%$ in
Financial.

\textbf{Without the lens (Cell A, raw cosine), a trap reaches the top-10
for 82--100\,\% of queries at $n = 5\,000$} in all four domains
($\HN = 0.8127$ Legal, $0.9967$ Medical, $0.9967$ Patent, $0.9900$
Financial), and for 81--100\,\% at $50\text{k}$ and $100\text{k}$ (Legal
$0.8161$ and $0.8161$, Medical and Patent $0.9967$, Financial $0.9867$;
\texttt{results/raw\_cosine/}). Legal's slightly lower value at $5\text{k}$ is
one trap that ties at rank 10 with identical clauses of other cases; the exact
search's sort breaks the tie differently in that pool. The lens
closes this gap to $0.0000$ across
all twelve $(\text{domain}, \text{pool})$ configurations.

Summary: for the declared cases, \textbf{twelve out of twelve}
production-scale measurements at $\HN = 0.0000$ among up to
$100\,000$ unseen documents, with $\HitC \in [0.93, 0.99]$ across every
configuration.

\begin{keyfindingbox}
For the declared cases, twelve out of twelve production-scale
(\textit{domain}, \textit{pool}) configurations reach $\HN = 0.0000$
among up to $100\,000$ unseen documents. Without the lens, a trap reaches
the top-10 for 82--100\,\% of queries at $n = 5\,000$. With the lens,
$\HN = 0.0000$ in all twelve configurations, with
$\HitC \in [0.93, 0.99]$.
\end{keyfindingbox}

\subsection{Cell D --- the served pipeline: the safety property at production scale}
\label{sec:celld}

Table~\ref{tab:results} measures the lens under exact nearest-neighbour
search (Cell~C): the utility ceiling, fully reproducible.
Table~\ref{tab:celld} reports Cell~D---the same lenses served by the
spectral accelerator of \S\ref{sec:accel}. The accelerator reaches the
$100\,000$-document pools, where a dense spectral build does not. $\HN$ is
unchanged ($0.0000$ at every scale); at the largest pools its $\HitC$ sits
a little below Cell~C's exact-search value, discussed after the table.

\begin{table}[h]
\centering
\small
\caption{Cell~D: the lens served by the spectral accelerator.
$\HN$ and $\HitC$ at $k = 10$, and the retrieval wall-clock, for four domains
$\times$ three pool sizes. The wall-clock is the measured elapsed time for the
retrieval of the full ${\sim}300$-query suite on a laptop CPU (the corpus is
pre-embedded; query embedding is excluded); the per-query figure is this
divided by the number of queries --- e.g.\ Legal $100\text{k}$ $0.62\,\text{s}
/ 299 \approx 2\,\text{ms}$ --- a suite average, not an isolated single-query
latency. Measured and SHA-anchored (Appendix~\ref{app:sha-anchors}); not part
of \pkg{}. Sources: sealed internal runs, one per domain (Legal, Medical,
Patent, Financial).}
\label{tab:celld}
\begin{tabular}{lrccr}
\toprule
\textbf{Domain} & $n$ & $\HN$ & $\HitC$ & \textbf{Suite wall} \\
\midrule
Legal     & 5\,897   & \textbf{0.0000} & 0.9799 & 0.03\,s \\
Legal     & 50\,897  & \textbf{0.0000} & 0.9866 & 0.26\,s \\
Legal     & 100\,929 & \textbf{0.0000} & 0.9498 & 0.62\,s \\
Medical   & 5\,300   & \textbf{0.0000} & 0.9733 & 0.05\,s \\
Medical   & 50\,300  & \textbf{0.0000} & 0.9700 & 0.45\,s \\
Medical   & 100\,300 & \textbf{0.0000} & 0.9633 & 1.06\,s \\
Patent    & 5\,300   & \textbf{0.0000} & 0.9701 & 0.05\,s \\
Patent    & 50\,300  & \textbf{0.0000} & 0.9535 & 0.43\,s \\
Patent    & 100\,300 & \textbf{0.0000} & 0.9269 & 1.03\,s \\
Financial & 5\,300   & \textbf{0.0000} & 0.9601 & 0.05\,s \\
Financial & 50\,300  & \textbf{0.0000} & 0.9435 & 0.48\,s \\
Financial & 100\,300 & \textbf{0.0000} & 0.9203 & 1.05\,s \\
\bottomrule
\end{tabular}
\end{table}

Across the four domains Cell~D holds $\HN = 0.0000$ at every pool size,
retains $\HitC \in [0.92, 0.99]$, and retrieves the full ${\sim}300$-query
suite over a $100\,000$-document pool in $0.6$--$1.1$\,s on a laptop CPU --- a
per-query average of ${\sim}2$--$4$\,ms, not an isolated single-query latency. The sealed same-pool runtime comparison
is Cell~D (accelerator) against the lens-only 3-stage cascade (Cell~B, exact
scoring over the pool), both on the \emph{identical} Legal dedup pool
(Table~\ref{tab:samepool}): the accelerator is $116$--$143\times$ faster, and
on that pool its $\HN$ is $0.0000$ --- better than the cascade's $0.08$--$0.10$
--- while its $\HitC$ sits $1.0$, $0.3$ and $4.0$ points below the cascade's
exact scoring at $5\text{k}$, $51\text{k}$ and $100\text{k}$. This is
\emph{not} a factor against the original dense SpectRAG, nor against the
\emph{open} full-pool exact nearest-neighbour backend (Cell~C), whose timing
is not sealed; it is Cell~B vs.\ Cell~D, one pool each.

\begin{table}[h]\centering\small
\caption{Sealed same-pool runtime (Legal, dedup pools): the
lens-only 3-stage cascade (Cell~B, exact scoring) and the spectral accelerator
(Cell~D) on the \emph{identical} pool. Wall-clock is the retrieval of the whole
299-case suite (per-query average = suite $/\,299$).
The accelerator is $116$--$143\times$ faster and holds $\HN = 0.0000$; its
$\HitC$ cost vs.\ the cascade's exact scoring is $1$--$4$ points, growing with
pool size.}
\label{tab:samepool}
\begin{tabular}{rlrcc}
\toprule
pool $n$ & backend & wall-clock & $\HN$ & $\HitC$ \\
\midrule
5\,897   & Cell~B cascade (exact)   & 3.85\,s  & 0.0836 & 0.9900 \\
5\,897   & Cell~D accelerator       & 0.027\,s & \textbf{0.0000} & 0.9799 \\
50\,897  & Cell~B cascade (exact)   & 35.33\,s & 0.0970 & 0.9900 \\
50\,897  & Cell~D accelerator       & 0.261\,s & \textbf{0.0000} & 0.9866 \\
100\,929 & Cell~B cascade (exact)   & 72.04\,s & 0.0970 & 0.9900 \\
100\,929 & Cell~D accelerator       & 0.620\,s & \textbf{0.0000} & 0.9498 \\
\bottomrule
\end{tabular}
\end{table}

At the largest pools Cell~D's $\HitC$ sits below Cell~C's exact-search
value (e.g.\ Legal $0.9833 \to 0.9498$ at $100\text{k}$). We do not
isolate the cause: Cell~C is measured on the open, re-baselined pools of
\pkg{} and Cell~D on the sealed original pools; the two differ in build
(and, for three domains, in size), so the Cell-C-vs-D difference mixes the
accelerator's approximate interior partitioning with the different pool
construction. No sealed same-pool comparison against the \emph{open full-pool
exact-NN} backend (Cell~C) exists, and we do not run one here; the sealed
same-pool numbers in Table~\ref{tab:samepool} are the cascade (Cell~B), not
Cell~C. An optional citation-fusion stage recovers the
Legal value to $0.9833$ at $100\text{k}$ and is a reserved component, not
evaluated here. Every Cell~D number is measured and SHA-anchored; the
accelerator is not part of \pkg{}, and the $\HN = 0.0000$ safety result
reproduces openly as Cell~C.

\begin{keyfindingbox}
Served by the spectral accelerator (Cell~D), the lens holds
$\HN = 0.0000$ at $n = 100\,000$ in all four domains, retrieving the
${\sim}300$-query suite in $0.6$--$1.1$\,s (${\sim}2$--$4$\,ms per query on
average), and $116$--$143\times$ faster than the lens-only 3-stage cascade
(Cell~B) on the same Legal pool (Table~\ref{tab:samepool}), where a dense
spectral build does not reach these pool sizes. The safety property is
identical to the exact-search Cell~C and reproduces openly; the latency is
the accelerator's measured, SHA-anchored contribution.
\end{keyfindingbox}

\subsection{\texorpdfstring{Complete $4\times 4$ cross-domain transfer matrix}{Complete 4x4 cross-domain transfer matrix}}
\label{sec:matrix}

\begin{figure}[h]
\centering
\includegraphics[width=0.85\textwidth]{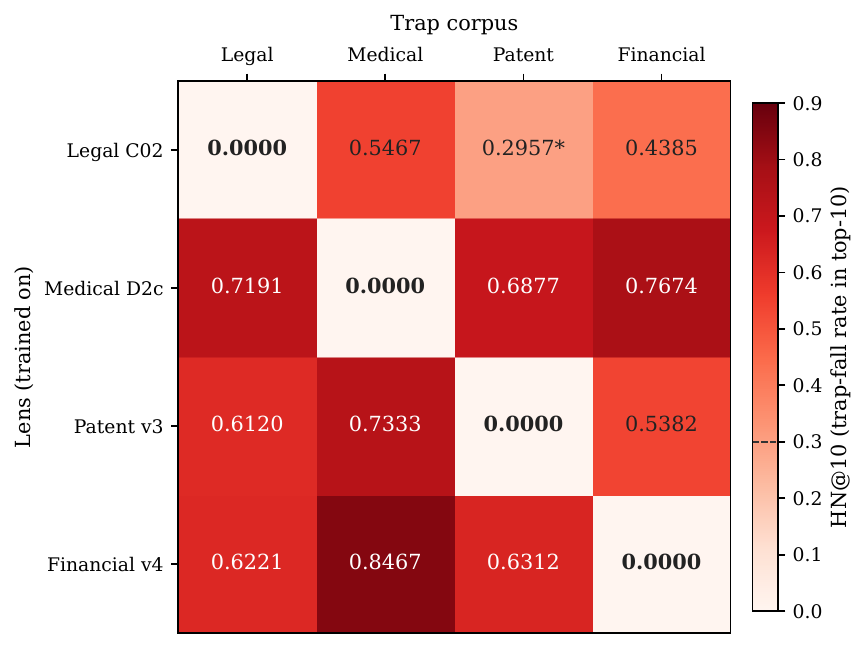}
\caption{Cross-domain transfer matrix at $n = 5\,000$. Diagonal cells
(matched-domain) reach $\HN = 0.0000$; eleven of twelve off-diagonal
cells stay in the $[0.44, 0.85]$ band; the twelfth (Legal C02 lens on
Patent traps, marked ${}^{\ast}$) sits at $\HN = 0.2957$, below the
$0.30$ the pre-registration had set (dashed marker on the colorbar),
and is reported as missed.}
\label{fig:matrix}
\end{figure}

Every lens applied to every domain's trap corpus at $n = 5\,000$, all
twelve off-diagonal cells now measured. Diagonal cells are
matched-domain (expected $\HN = 0.0000$). Off-diagonal cells apply a
lens to another domain's cases; the pre-registration had set
$\HN \geq 0.30$ for them. All twelve off-diagonal cells were
measured through the same \texttt{miss\_diagnostic.py} code path that
produced the four on-diagonal $0.0000$ results, so all sixteen
numbers are apples-to-apples comparable. Figure~\ref{fig:matrix} plots
the matrix as a heatmap; the numeric values follow in
Table~\ref{tab:matrix}.

\begin{table}[h]
\centering
\small
\caption{Cross-domain transfer matrix, numeric values.
Corresponds to the heatmap in Figure~\ref{fig:matrix}.}
\label{tab:matrix}
\begin{tabular}{lcccc}
\toprule
\textbf{Lens $\downarrow$ / Trap corpus $\to$} & Legal & Medical & Patent & Financial \\
\midrule
\textbf{Legal C02}   ($d\!=\!11$) & \textbf{0.0000} & 0.5467 & $0.2957^{\ast}$ & 0.4385 \\
\textbf{Medical D2c} ($d\!=\!32$) & 0.7191          & \textbf{0.0000} & 0.6877 & 0.7674 \\
\textbf{Patent v3}   ($d\!=\!32$) & 0.6120          & 0.7333 & \textbf{0.0000} & 0.5382 \\
\textbf{Financial v4}($d\!=\!32$) & 0.6221          & 0.8467 & 0.6312 & \textbf{0.0000} \\
\bottomrule
\end{tabular}
\end{table}

\begin{itemize}
  \item \textbf{Diagonal}: 4\,/\,4 cells at $\HN = 0.0000$.
  \item \textbf{Off-diagonal, at or above the pre-registered $0.30$}:
    11\,/\,12 cells, $\HN \in [0.44, 0.85]$, mean ${\approx}0.65$.
  \item \textbf{Off-diagonal, below it}: 1 cell (marked ${}^{\ast}$),
    the Legal C02 lens on the Patent trap corpus, $\HN = 0.2957$. The
    pre-registered criterion is missed in this cell.
\end{itemize}

The criterion is missed by $0.0043$, and we report it as missed. One
plausible reading, not a test: the Legal lens ($d = 11$, the narrowest
in the panel, trained on the legal taxonomy) is closer to the
Patent taxonomy (claim-scope reversals inside legalistic clauses) than
to the medical or financial ones. \S\ref{sec:ablation} reports a
follow-up retrain. The lens does reduce the Patent traps, from raw
cosine's $0.9967$ to $0.2957$, but it still leaves ${\sim}89$ of the 301
in the top-10.

Every off-diagonal cell lies below the target domain's raw cosine
($0.82$--$1.00$ at this size): the lenses carry some structure the
domains share and transfer partly across them, but none reaches the
declared-case level off-domain.

\begin{keyfindingbox}
A lens reaches $\HN = 0.0000$ on the cases it was declared for.
Applied to another domain it transfers partly: every off-diagonal cell
is below raw cosine, eleven of twelve at $\HN \in [0.44, 0.85]$, and the
twelfth, $\HN = 0.2957$, is below the pre-registered $0.30$ and reported
as missed. None reaches the matched-domain $0.0000$.
\end{keyfindingbox}

\subsection{\texorpdfstring{Follow-up --- the Legal lens retrained at $d=32$}{Follow-up --- the Legal lens retrained at d=32}}
\label{sec:ablation}

\begin{figure}[H]
\centering
\includegraphics[width=0.92\textwidth]{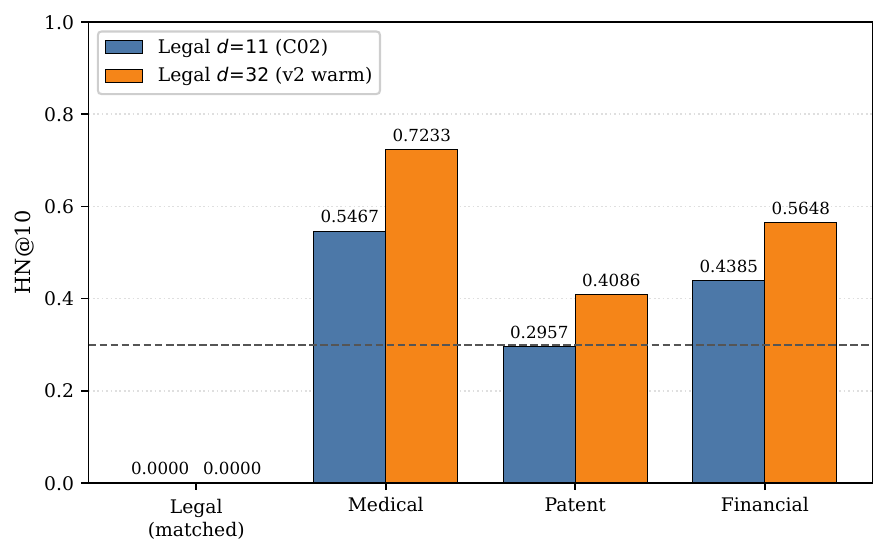}
\caption{The Legal lens retrained at $d = 32$. The dashed horizontal
line marks the $0.30$ the pre-registration had set. Cross-domain $\HN$ rises \emph{uniformly}
by $+0.11$ to $+0.16$ across all three foreign corpora when the Legal
lens is retrained at $d = 32$ instead of $d = 11$ (matching the other
three domain lenses). The rise is not selective on Patent, which fits
a capacity reading better than a Legal--Patent affinity; with one seed
and a different training history it is not a proof. Both the $d = 11$ and
$d = 32$ Legal lenses hold matched-domain $\HN = 0.0000$.}
\label{fig:ablation}
\end{figure}

Motivated by the one off-diagonal cell below the pre-registered $0.30$ in
\S\ref{sec:matrix} (Legal C02 at $d = 11$ applied to Patent traps,
$\HN = 0.2957$), we retrained the Legal lens at $d = 32$ with
the exact hyperparameter set used for Medical D2c / Patent v3 /
Financial v4 ($\beta = 15$, $\text{margin} = 0.10$, $\text{seed} = 42$,
cold-start followed by a warm-start round). The training history is
not that of the $d = 11$ lens (C02 has a longer warm-start lineage) and
we ran one seed, so this does not isolate the dimension as the only
variable; it is a follow-up observation. Figure~\ref{fig:ablation} plots the results at
$n = 5\,000$; numeric values follow in Table~\ref{tab:ablation}.

\begin{table}[h]
\centering
\small
\caption{The Legal lens at $d = 11$ and $d = 32$, numeric values. Corresponds to the bars
in Figure~\ref{fig:ablation}.}
\label{tab:ablation}
\begin{tabular}{lccc}
\toprule
\textbf{Corpus} & \textbf{Legal C02} ($d\!=\!11$) $\HN$ & \textbf{Legal $d\!=\!32$ v2} $\HN$ & $\Delta$ \\
\midrule
Legal (matched) & 0.0000 & 0.0000 & $+0.0000$ \\
Medical         & 0.5467 & 0.7233 & $+0.1766$ \\
Patent          & 0.2957 & \textbf{0.4086} & $+0.1129$ \\
Financial       & 0.4385 & 0.5648 & $+0.1263$ \\
\bottomrule
\end{tabular}
\end{table}

Three things to notice. First, both lenses hold matched-domain
$\HN = 0.0000$, so the $d = 32$ retrain adds cross-domain separation at no
matched-domain cost. Second, on every foreign corpus the cross-domain
$\HN$ rises uniformly by $0.11$ to $0.18$, not
selectively on Patent. Third, the rank order across the three foreign
corpora ($\text{Medical} > \text{Financial} > \text{Patent}$) is
preserved between $d = 11$ and $d = 32$---Patent stays the softest
cross-domain target for the Legal lens, now above the $0.30$ the
pre-registration had set.

The uniform rise fits a capacity reading better than a Legal--Patent
affinity. With one seed and a different training history this is an
explanation, not a test: the pre-registered criterion for the $d = 11$
lens stays missed.

The $d = 32$ Legal lens is retained as an alternative configuration
and archived alongside C02 in Appendix~\ref{app:sha-anchors}. Both hold
matched-domain $\HN = 0.0000$; the $d = 32$ lens additionally lifts the
one sub-threshold cross-domain cell above $0.30$. Which to deploy in a
Legal-domain production stack is an operator choice, not a mechanism
question.

\begin{keyfindingbox}
Retrained at $d = 32$, the Legal lens moves the Legal $\to$ Patent cell
from $\HN = 0.2957$ to $\HN = 0.4086$, with $+0.11$ to $+0.18$ on all
three foreign corpora. This explains the missed cell; it does not
retroactively pass the pre-registered criterion, and one seed with a
different training history does not isolate the dimension.
\end{keyfindingbox}

\subsection{Wall-clock at production scale}
\label{sec:wallclock}

CPU-only single-core measurements on M4 (10-core Apple Silicon, no
GPU). Cell~D (the spectral accelerator) at $n = 100\,300$ pool:

\begin{figure}[h]
\centering
\includegraphics[width=0.92\textwidth]{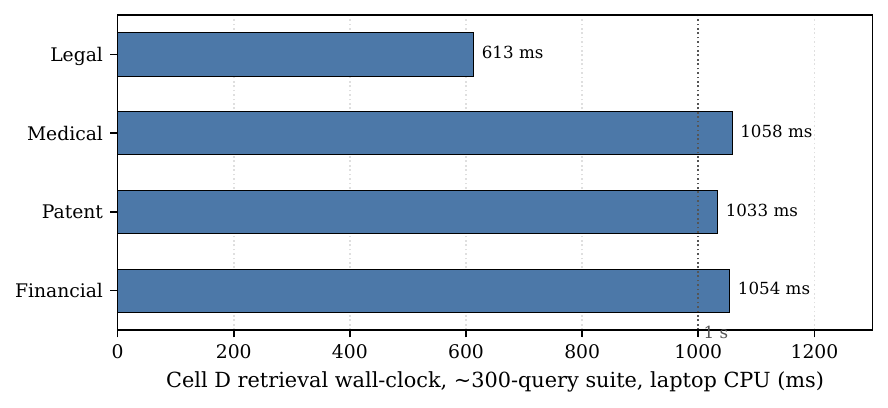}
\caption{Cell~D retrieval wall-clock for the ${\sim}300$-query suite at
$n = 100\,300$ on a laptop CPU. All four domains run the whole suite in under
$1.1$\,s (a per-query average of ${\sim}2$--$4$\,ms). The lens itself is a
single $1024 \times d$ matmul (${\sim}1$\,ms); the rest is the downstream
accelerator (Cell~D). One pool size is shown; it makes no claim about how the
time grows with $n$.}
\label{fig:wallclock}
\end{figure}

The lens projection itself is a single $1024 \times d$ matrix
multiply per query, so the wall-clock is dominated by the downstream
retriever. All four domains run the ${\sim}300$-query retrieval suite in
$0.6$--$1.1$\,s at $n = 100\,000$ on a laptop CPU --- a per-query average of
${\sim}2$--$4$\,ms. All reported wall-clocks are CPU; we claim no GPU figure.

\paragraph{Reproducibility note.} The wall-clock numbers above assume
the proprietary accelerator (Cell~D). Readers reproducing the paper with the
open-source substitute (exact nearest-neighbour search on the projected
$d$-dimensional pool) obtain $\HN = 0.0000$ on all twelve configurations
of \S\ref{sec:results}; its time per query was not measured beside
Figure~\ref{fig:wallclock}. Both backends we tested---the accelerator and
the open exact-nearest-neighbour substitute---return zero observed trap hits
on their respective pools; we do not claim independence from an arbitrary
retriever (the observation is limited to the two backends compared).

\subsection{Baselines: random projection and PCA at the same dimension}
\label{sec:baselines}
Is the lens's $\HN = 0.0000$ merely a consequence of reducing to $d$? Two
trap-agnostic projections at the same output dimension say no. A random
Gaussian projection (seed 42) and unsupervised PCA (fit on the distractors
only, no label leakage), at $d \in \{11, 32\}$, were measured through the
\emph{same} exact-nearest-neighbour code as Cell~C (open,
reproducible, SHA-anchored). All sixteen (baseline, domain) cells leave the
trap in the top-10 for far more than the pre-registered $0.30$:

\begin{table}[h]\centering\small
\caption{Trap-fall rate $\HN$ of trap-agnostic projections at the shipped
dimensions, versus the trained lens, on the four matched-domain $5\text{k}$
pools (exact search).}
\label{tab:baselines}
\begin{tabular}{lcccc}
\toprule
projection & Legal & Medical & Patent & Financial \\
\midrule
random projection, $d{=}11$ & 0.5251 & 0.7000 & 0.6478 & 0.6478 \\
random projection, $d{=}32$ & 0.7659 & 0.9600 & 0.9668 & 0.9236 \\
PCA, $d{=}11$ & 0.5418 & 0.8167 & 0.8472 & 0.5748 \\
PCA, $d{=}32$ & 0.6020 & 0.9367 & 0.9668 & 0.6445 \\
\textbf{trained lens} & \textbf{0.0000} & \textbf{0.0000} & \textbf{0.0000} & \textbf{0.0000} \\
\bottomrule
\end{tabular}
\end{table}

\noindent
Reduction to $d$ alone leaves the trap in the top-10 for $53$--$97\,\%$ of the
declared cases; the lens reaches $0.0000$. The trap immunity is a property of
what the lens learned, not of the projection to $d$.

\subsection{Robustness to rewording (Legal)}
\label{sec:robustness}
The declared cases can be stated in words the lens was not trained on. We
rewrite the declared Legal cases with a fixed, fully-listed carrier lexicon
under a per-case check that keeps each case's trap-discriminating tokens
byte-identical, so the one-token trap edit is preserved by construction; the check proves preserved token \emph{differences}, not full semantic
equivalence. Two tests measure different things.

\paragraph{Joint rewording (query and corpus together).} Applying the lexicon
identically to $A$, $B$ and $C$ rewrites the query \emph{and} the stored
documents consistently; $194$ of the $299$ cases admit such a safe rewrite. The
rewordings are faithful (mean E5 cosine to the source $0.99$) but deliberately
mild --- the safe lexicon never touches a discriminator, so this is a lower
bound on the stress. Here the shipped lens holds $\HitC = 0.9845$ and
$\HN = 0.0155$ ($3$ of $194$ cases), against $0.0052$ ($1$ of $194$) on the
trained phrasings; without the lens the trap reaches the top-10 for about
$81\,\%$ of these cases. Declaring the phrasings closes the residual: an
improved lens fine-tuned from the shipped one on the declared cases plus one
reworded variant reaches $\HN = 0.0000$ on the declared phrasings and $0.0103$
on a second, held-out variant (vs.\ $0.0258$ for the shipped lens), at about
one point of $\HitC$ on the originals --- the concept-scoped method by design,
not a post-hoc fit to the evaluation.

\paragraph{Query-only rewording (the practical case).} Here the query vector
is the reworded $A'$ but the search pool is the \emph{unchanged} original
corpus --- the original $A$, $B$, $C$ and distractors, with no reworded row in
it (the case's own original $A$ is still excluded from its ranking). This asks
whether a differently-worded user query still finds the right passage in an
unchanged corpus. The \emph{safety} property holds --- $\HN = 0.0052$ ($1$ of
$194$) --- but \emph{utility} drops: $\HitC = 0.7938$ for the shipped lens, down
from $0.9845$ under joint rewording, because the reworded query matches the
\emph{original} paraphrase $C$ less well. The improved lens scores $0.8763$ on
these same queries, \emph{but this phrasing (v1) is in its training set}, so
that figure is fit to declared wordings, not generalization: on a phrasing it
was \emph{not} trained on (v2), the two lenses are comparable --- shipped
$0.8093$, improved $0.7784$ --- i.e.\ declaring and fitting helps the declared
wordings, not arbitrary unseen queries. So the lens's trap suppression is robust
to query rewording, but its retrieval utility on reworded queries is markedly
lower (${\sim}0.78$--$0.81$) and does not improve for undeclared phrasings.
These absolute counts are small and the observation is correspondingly limited.

\section{Discussion}
\label{sec:discussion}

\subsection{Interpretation of the results}

\begin{itemize}
  \item \textbf{Shared structure, partial transfer.} The $4\times 4$
    matrix (\S\ref{sec:matrix}) does not show a purely domain-specific
    lens. Every foreign lens lowers $\HN$ below raw cosine---the Legal
    lens takes Patent from $0.9967$ to $0.2957$---so the lenses carry
    some structure the domains share. None reaches the declared-case
    $0.0000$ off-domain: full suppression holds for the declared cases
    only.
  \item \textbf{``It's a per-corpus artefact.''} Four independent
    corpora, four independent trap taxonomies, four independent
    distractor sources---same result. Reproducing this on a fifth
    independent domain (bioinformatics, semiconductor process,
    insurance-policy, or scientific-claim traps) with the same recipe
    would push the ``not artefact'' claim past reasonable doubt; this
    is our next experimental milestone.
\end{itemize}

\subsection{What the evidence does not establish}
\label{sec:limits}

\begin{limitationsbox}
\begin{enumerate}
  \item \textbf{Constructed traps.} The 299--301 test cases per domain
    are hand-authored matched-shape traps---single-token in three domains,
    short edits in legal. This is a
    deliberate stress test, not a natural-language corpus. Transfer to
    naturally-occurring adversarial examples (multi-token flips,
    long-form contradictions) is not established here.
  \item \textbf{Single encoder tested.} All reported numbers use
    E5-Large-v2. The training procedure is architecture-agnostic in
    principle; behaviour on BGE-Large / GTE-Large is a hypothesis for
    further experiments, not measured.
  \item \textbf{Outside the declared set, by design.} Cases outside a
    lens's declared set---including new statement pairs of the same
    type---are not covered (\S\ref{sec:declared}); one declares and
    trains a lens for them, or combines lenses.
  \item \textbf{No full transfer to a new domain.} A foreign lens
    lowers $\HN$ but does not reach $0.0000$ (\S\ref{sec:matrix}); a new
    domain's cases need their own lens, trained on ${\sim}300$
    in-domain triples.
  \item \textbf{Cell D latency is proprietary.} The $0.6$--$1.1$\,s suite
    retrieval (Cell~D; ${\sim}2$--$4$\,ms per query) relies on our commercial spectral
    accelerator (\S\ref{sec:accel}); its numbers are measured and
    SHA-anchored, but the implementation is not released. The open-source
    exact-nearest-neighbour substitute (Cell~C) reproduces every $\HN$
    number; its time per query was not measured here.
  \item \textbf{Other components out of scope.} A proprietary
    enterprise retriever and additional re-ranking and fusion components
    are out of scope and not part of this release; Cell~B's retrieval quality
    is tabulated only on Legal, in the same-pool comparison
    (Table~\ref{tab:samepool}); the other domains' cascade quality is not,
    to keep the claim isolated to the lens projection.
  \item \textbf{No E5 task prefixes.} We embed without the
    \texttt{query:}\,/\,\texttt{passage:} prefixes that the E5 model
    card recommends; results with the recommended prefixes are not
    measured.
  \item \textbf{Some baselines still open.} The projection without its
    preservation or hinge term, a cross-encoder reranker, an evaluation on
    NevIR \citep{weller2024nevir}, and several seeds for the $d = 32$
    follow-up are not part of this paper. Random projection and PCA at the
    same $d$ (\S\ref{sec:baselines}) and the cascade-versus-accelerator timing
    on the same pool (\S\ref{sec:celld}) \emph{are} now measured.
  \item \textbf{Determinism in the tested environment.} The
    bit-identical re-runs hold for the pinned dependencies on the CPUs
    we tested; other hardware or library versions may differ in the
    last digits of a float.
\end{enumerate}
\end{limitationsbox}

The Limitations box above lists the points a critical reader should
flag; a few of them warrant longer discussion.

\paragraph{Constructedness is intentional.} The 299--301 test cases
per domain are hand-authored matched-shape traps (single-token in three
domains, short multi-token edits in legal)---adversarial constructions,
not sampled from a natural distribution of retrieval failures. This is
deliberate: the paper's claim is that a specific stress-test class (small
meaning-reversing edits inside short clauses of the target-domain
vocabulary) is fully suppressible with a small learned projection. Natural-language adversarial transfer
(multi-token flips, long-form contradictions, real user queries that
misfire) is a separate empirical question and is future work. We
prefer the stress-test framing because it is falsifiable: any single
uncured test case with the trap in top-10 counts as a failure of the
claim.

\paragraph{Retrieval-utility ceiling.} Under exact search (Cell~C)
$\HitC$ ranges from $0.93$ to $0.99$ across the domains; under the
accelerator (Cell~D) it is $0.92$ to $0.99$. On the sealed dedup pools, where
the lens-only 3-stage cascade (Cell~B, exact scoring) and the accelerator
(Cell~D) ran on the \emph{identical} pool (Table~\ref{tab:samepool}), the accelerator's cost versus the cascade's exact
scoring is $-1.0$, $-0.3$ and $-4.0$ percentage points of $\HitC$ at
$5\text{k}$, $51\text{k}$ and $100\text{k}$ --- small, growing with pool size
(the approximate interior partitioning) --- while it holds $\HN = 0.0000$ and
runs $116$--$143\times$ faster than the cascade on that pool ($0.62$\,s vs.\
$72$\,s at $100\text{k}$). The \emph{open} full-pool exact-NN backend
(Cell~C) was not timed on that pool (\S\ref{sec:celld}). This is operationally
useful but not perfect retrieval. The remaining misses are typically in the
``$C$-paraphrase differs too much from $A$'' pattern --- a data-quality
issue in the trap corpus that we address with miss-driven curation
(documented in the reproducibility repository as the
\texttt{curate\_*\_paraphrases.py} scripts).

\paragraph{Retrieval speed as a contribution.} The $0.6$--$1.1$\,s suite
retrieval wall-clock at $n = 100\,000$ (Cell~D, \S\ref{sec:celld}) is the
accelerator's contribution: the boundary-conditioned equilibrium solve of
\S\ref{sec:accel} serves the lens where a dense spectral \emph{build} does
not reach the pool at all, and---on the same Legal pool---$116$--$143\times$
faster than the lens-only 3-stage cascade (Table~\ref{tab:samepool}). These
numbers are measured and
SHA-anchored; the accelerator's implementation is reserved for commercial
use, and an exact-nearest-neighbour substitute (Cell~C) reproduces every
$\HN$ number as the open reproducibility path. The safety claim does not
depend on the accelerator; the scale-and-latency claim does.

\subsection{Limits of the transfer experiment}

Four domains and twelve mismatched pairs are a finite sample. The matrix
shows where each lens applies; it does not prove that the recipe cannot
fail in another domain. Our best hypothesis for where it might: domains
whose traps do not admit clean small edits, or where the $A/B/C$
shape is hard to keep (long-form narrative, multi-paragraph reasoning).
The fifth-domain candidates in \S\ref{sec:ongoing} are chosen partly to
probe this. For the Legal $\to$ Patent cell, \S\ref{sec:ablation}
reports the $d = 32$ follow-up, which lifts that cell above $0.30$ while
both the $d = 11$ and $d = 32$ lenses keep matched-domain $\HN = 0.0000$.

\subsection{Reproducibility protocol summary}

Any reader with 24\,GB of RAM, a modern multi-core CPU, and access to
a 1024-dimensional dense encoder can reproduce the claim within a
working day:

\begin{enumerate}
  \item Construct 300 matched-shape $(A, B, C)$ triples in the target
    domain (2--3 hours of deterministic-script authoring; template
    provided).
  \item Build a sentence-split distractor pool from a domain corpus
    (target $\geq 50\,000$ unique 8--30-word clauses).
  \item Embed with E5-Large-v2, no prefix, $L_2$-normalise.
  \item Run the preflight duplicate check
    (\texttt{preflight\_dedup\_check.py}). Abort if $\geq 5\,\%$
    duplicates.
  \item Train the safety lens at $d = 24$ or $32$, $\beta = 15$ or
    $17$, $\text{margin} = 0.10$, $\text{seed} = 42$
    (\texttt{steering\_lens\_train.py}). Warm-start a second round if
    $\HN$ is not yet $0.0000$.
  \item Measure Cell C at pool sizes $\{5\text{k}, 50\text{k},
    100\text{k}\}$. Optionally apply the lens to two other domains' trap
    corpora, to see where it applies.
\end{enumerate}

Wall-clock: approximately one working day end-to-end on a single
CPU server.

\section{Deployment footprint: small, specialised, hardened indexes}
\label{sec:deployment}

A deployed application stores only the lens-projected vectors, not the
$1024$-D embeddings. For the Legal lens that is $d = 11$ floats per document
--- $44$ bytes against $4096$, a $93\times$ smaller index; the $d = 32$ domains
are $32\times$ smaller. Exact nearest-neighbour cost scales with the dimension,
so search in the projected space is proportionally cheaper, and at these
dimensions an exact scan is realistic again rather than an approximate index.
On the cases the lens is declared for, this smaller index is also \emph{safer}
than raw-cosine search over the full embedding ($\HN = 0.0000$ versus
$0.81$--$1.00$); where the raw encoder is weak it is also \emph{more accurate}
(Legal $\HitC$ $0.99$ vs.\ $0.67$). The projected store is application-specific
by construction: one declared lens and one index per scope, and the $1024$-D
embedding cannot be recovered from it.

\begin{figure}[t]
\centering
\begin{tikzpicture}[node distance=2.6mm,
  bigidx/.style={rectangle, rounded corners=3pt, draw=pelBorder, line width=0.7pt,
    fill=pelLight, text width=4.0cm, align=center, inner sep=7pt, font=\small},
  smallidx/.style={rectangle, rounded corners=2pt, draw=pelBorder, line width=0.5pt,
    fill=pelBlue!8, text width=4.4cm, align=left, inner sep=4pt, font=\scriptsize},
  htag/.style={font=\small\bfseries\sffamily, text width=4.4cm, align=center}]

  \node[htag] (lt) {One generalist index};
  \node[bigidx, below=2mm of lt] (big)
    {\textbf{1024-D} general-purpose\\ vector database\\[5pt]
     $4096$\,B / vector\\ one shared store\\[5pt]
     $\HN$ up to $1.00$\\ (semantic-reversal traps)\\ costly to shard};

  \node[htag, right=3.2cm of lt] (rt) {Many specialised indexes \textit{(this work)}};
  \node[smallidx, below=2mm of rt] (s1) {\textbf{Legal} --- 11-D lens index --- $44$\,B/vec};
  \node[smallidx, below=of s1] (s2) {\textbf{Medical} --- 32-D --- $128$\,B/vec};
  \node[smallidx, below=of s2] (s3) {\textbf{Contract} --- 11-D --- $44$\,B/vec};
  \node[smallidx, below=of s3] (s4) {\textbf{Internal-policy} --- 11-D --- $44$\,B/vec};
  \node[font=\scriptsize, text width=4.4cm, align=center, below=1.8mm of s4] (props)
    {\textit{$\vdots$ one declared lens + index per scope}\\[2pt]
     $\HN = 0$ on declared cases $\;\cdot\;$ ${\sim}90\times$ smaller $\;\cdot\;$
     dimension-proportional search $\;\cdot\;$ isolatable per tenant};

  \draw[flowarrow] (big.east) -- (s2.west)
    node[midway, above, font=\scriptsize\sffamily]{declare $+$ fit}
    node[midway, below, font=\scriptsize\sffamily]{one small lens / scope};
\end{tikzpicture}
\caption{Deployment footprint. The lens's output \emph{is} the stored
representation: a few floats per document instead of $1024$. A specialised,
semantic-reversal-hardened index per domain or tenant ($\HN = 0$ on its
declared cases, ${\sim}90\times$ smaller, dimension-proportional search) becomes
a practical alternative to one large general-purpose embedding store. The
pattern is enabled by the measured footprint; the comparison is architectural,
not a benchmark.}
\label{fig:deployment}
\end{figure}
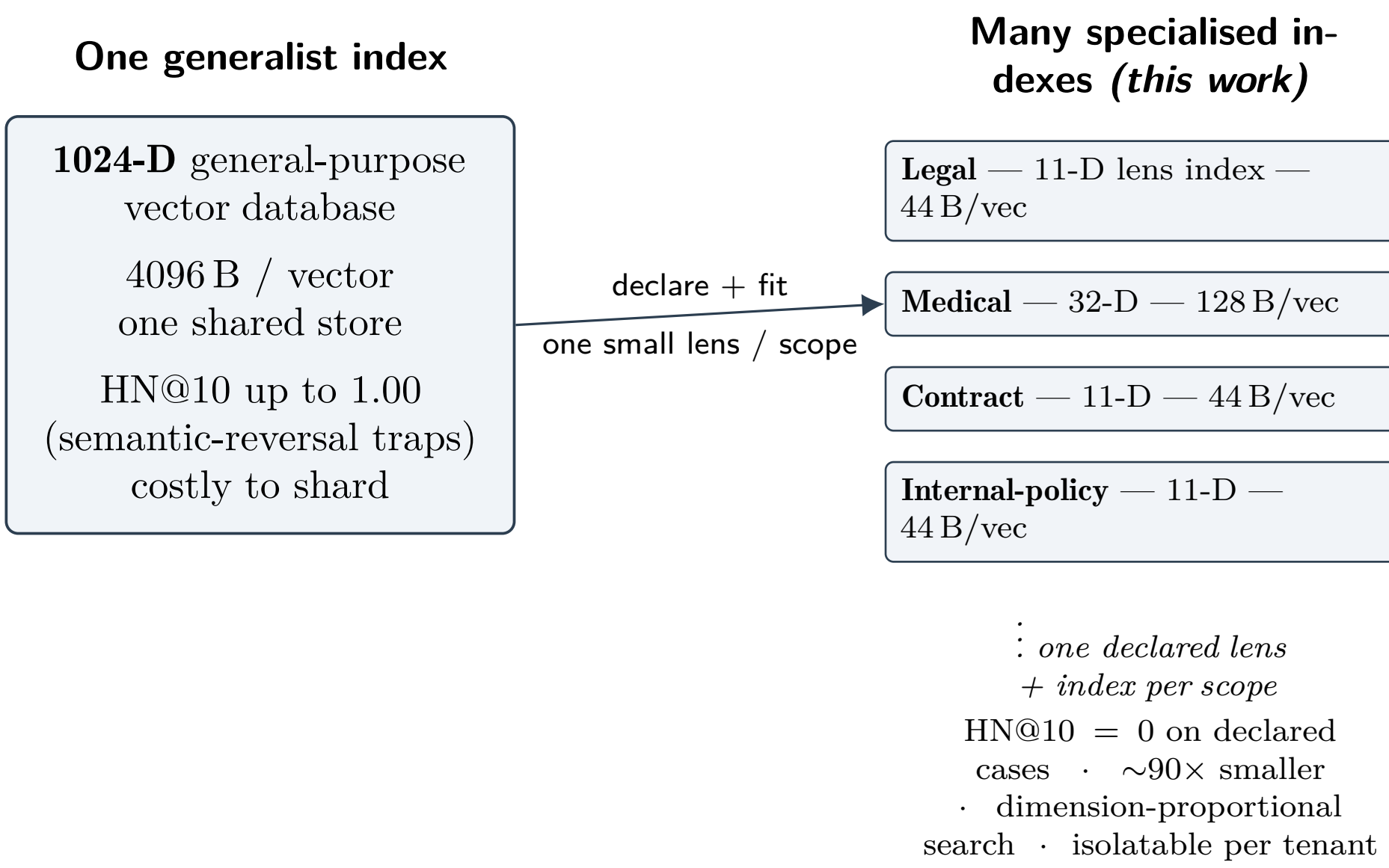

\emph{Implication --- a deployment pattern, not a benchmark.} Because a
specialised, semantic-reversal-hardened index becomes this small and this cheap
to search, running \emph{many} of them --- one per domain, tenant, or
compliance boundary, each with its own declared lens --- becomes a rational
alternative to a single large general-purpose vector database
(Figure~\ref{fig:deployment}): cheaper to store, faster to search, hardened per
scope, versionable, and trivially isolatable into separate processes or
tenants. It inverts the usual ``reach for a bigger generalist encoder''
reflex --- a tiny learned projection makes \emph{specialisation} the cheap
option. We state this as the architecture the measured footprint enables; a
cost or multi-store benchmark is future work. The query-only result
(\S\ref{sec:robustness}) bounds it: utility holds for the declared cases, not
for arbitrary undeclared query phrasings.

\section{Related Ongoing Work}
\label{sec:ongoing}

\paragraph{\pkg.} Reproducibility package, released
under AGPL-3.0 at \url{https://github.com/pelloran-inc/pelloran-safety-lens}. Contains the full 6-step recipe implementation,
the four domains' trap sets, deterministic generators, lens
trainer, measurement harness for Cell A (raw cosine) and the lens
(with exact nearest-neighbour search in place of the proprietary
accelerator of Cell D), duplicate-check guardrail, and the cross-domain
transfer tool. All four distractor pools rebuild from pinned public
sources (\texttt{DATA.md}): the medical pool from the CC0
\texttt{jmhb/pubmed\_bioasq\_2022}, the legal pool from LexGLUE and the
public-domain SEC EDGAR corpus. The model and dataset revisions and the
library versions are pinned, and the SHA-256 of every pool binary and
result ships with the package. Anyone may use, study, modify and
share the package under AGPL-3.0, whose terms require that the source
code of a modified version be offered to all its users, including users
who reach it over a network. A commercial licence without these terms,
and a licence for the proprietary accelerator, are available from
\texttt{opensource@pelloran.com}.

\paragraph{Domain expansion.} Fifth-domain candidates under active
discussion: bioinformatics protein-annotation traps, semiconductor
process-flow traps, insurance-claim/policy traps, and scientific-claim
traps. Each additional domain extends the transfer matrix in
\S\ref{sec:matrix}.

\paragraph{Encoder generalisation.} Reproducing the four-domain matrix
with BGE-Large and GTE-Large as the base encoder, to confirm that the
recipe is encoder-agnostic rather than E5-specific.

\paragraph{Pipeline composition.} Further re-ranking and fusion stages
on top of the projection are out of scope and not part of this release;
they are left to a future paper on pipeline composition rather than the
per-domain generalisation claim.

\bibliographystyle{plainnat}
\bibliography{whitepaper}

\begin{thebibliography}{18}
\providecommand{\natexlab}[1]{#1}
\providecommand{\url}[1]{\texttt{#1}}
\expandafter\ifx\csname urlstyle\endcsname\relax
  \providecommand{\doi}[1]{doi: #1}\else
  \providecommand{\doi}{doi: \begingroup \urlstyle{rm}\Url}\fi

\bibitem[Chen et~al.(2024)Chen, Xiao, Zhang, Luo, Lian, and Liu]{chen2024bge}
Jianlv Chen, Shitao Xiao, Peitian Zhang, Kun Luo, Defu Lian, and Zheng Liu.
\newblock {BGE M3-Embedding}: Multi-lingual, multi-functionality,
  multi-granularity text embeddings through self-knowledge distillation.
\newblock \emph{arXiv preprint arXiv:2402.03216}, 2024.
\newblock URL \url{https://arxiv.org/abs/2402.03216}.

\bibitem[Dodge et~al.(2021)Dodge, Sap, Marasovi\'{c}, Agnew, Ilharco,
  Groeneveld, Mitchell, and Gardner]{dodge2021c4}
Jesse Dodge, Maarten Sap, Ana Marasovi\'{c}, William Agnew, Gabriel Ilharco,
  Dirk Groeneveld, Margaret Mitchell, and Matt Gardner.
\newblock Documenting large webtext corpora: A case study on the colossal clean
  crawled corpus.
\newblock In \emph{Proceedings of EMNLP}, pages 1286--1305, 2021.

\bibitem[Guo et~al.(2019)Guo, Gardner, You, Wilson, and
  Weinberger]{guo2019blackbox}
Chuan Guo, Jacob Gardner, Yurong You, Andrew~Gordon Wilson, and Kilian
  Weinberger.
\newblock Simple black-box adversarial attacks.
\newblock In \emph{Proceedings of ICML}, pages 2484--2493, 2019.

\bibitem[Karpukhin et~al.(2020)Karpukhin, O\u{g}uz, Min, Lewis, Wu, Edunov,
  Chen, and Yih]{karpukhin2020dpr}
Vladimir Karpukhin, Barlas O\u{g}uz, Sewon Min, Patrick Lewis, Ledell Wu,
  Sergey Edunov, Danqi Chen, and Wen-tau Yih.
\newblock Dense passage retrieval for open-domain question answering.
\newblock In \emph{Proceedings of EMNLP}, pages 6769--6781, 2020.

\bibitem[Khattab and Zaharia(2020)]{khattab2020colbert}
Omar Khattab and Matei Zaharia.
\newblock {ColBERT}: Efficient and effective passage search via contextualized
  late interaction over {BERT}.
\newblock In \emph{Proceedings of SIGIR}, pages 39--48, 2020.

\bibitem[Lee et~al.(2022)Lee, Ippolito, Nystrom, Zhang, Eck, Callison-Burch,
  and Carlini]{lee2022dedup}
Katherine Lee, Daphne Ippolito, Andrew Nystrom, Chiyuan Zhang, Douglas Eck,
  Chris Callison-Burch, and Nicholas Carlini.
\newblock Deduplicating training data makes language models better.
\newblock In \emph{Proceedings of ACL}, pages 8424--8445, 2022.

\bibitem[Li et~al.(2023)Li, Zhang, Zhang, Long, Xie, and Zhang]{li2023gte}
Zehan Li, Xin Zhang, Yanzhao Zhang, Dingkun Long, Pengjun Xie, and Meishan
  Zhang.
\newblock Towards general text embeddings with multi-stage contrastive
  learning.
\newblock \emph{arXiv preprint arXiv:2308.03281}, 2023.
\newblock URL \url{https://arxiv.org/abs/2308.03281}.

\bibitem[Robertson and Zaragoza(2009)]{robertson2009bm25}
Stephen Robertson and Hugo Zaragoza.
\newblock The probabilistic relevance framework: {BM25} and beyond.
\newblock \emph{Foundations and Trends in Information Retrieval}, 3\penalty0
  (4):\penalty0 333--389, 2009.

\bibitem[Sahinoglu et~al.(2026)Sahinoglu, Mohammed, Kazangirler, and
  Kazangirler]{sahinoglu2026spectrag}
Zafer Sahinoglu, Abdulazeez Mohammed, Buse~Yaren Kazangirler, and Cahit~Berkay
  Kazangirler.
\newblock {SpectRAG}: Weighted chunk graphs for token-budgeted context
  selection in multi-hop retrieval-augmented generation.
\newblock \emph{arXiv preprint arXiv:2610.36991}, 2026.
\newblock URL \url{https://arxiv.org/abs/2610.36991}.

\bibitem[Schroff et~al.(2015)Schroff, Kalenichenko, and
  Philbin]{schroff2015facenet}
Florian Schroff, Dmitry Kalenichenko, and James Philbin.
\newblock {FaceNet}: A unified embedding for face recognition and clustering.
\newblock In \emph{Proceedings of CVPR}, pages 815--823, 2015.

\bibitem[Sohn(2016)]{sohn2016npair}
Kihyuk Sohn.
\newblock Improved deep metric learning with multi-class {N-pair} loss
  objective.
\newblock In \emph{Advances in Neural Information Processing Systems},
  volume~29, pages 1857--1865, 2016.

\bibitem[Song et~al.(2020)Song, Rush, and Shmatikov]{song2020adversarial}
Congzheng Song, Alexander~M. Rush, and Vitaly Shmatikov.
\newblock Adversarial semantic collisions.
\newblock In \emph{Proceedings of EMNLP}, pages 4198--4210, 2020.

\bibitem[Turner et~al.(2023)Turner, Thiergart, Udell, Leech, Mini, and
  MacDiarmid]{turner2023activation}
Alexander~Matt Turner, Lisa Thiergart, David Udell, Gavin Leech, Ulisse Mini,
  and Monte MacDiarmid.
\newblock Activation addition: Steering language models without optimization.
\newblock \emph{arXiv preprint arXiv:2308.10248}, 2023.
\newblock URL \url{https://arxiv.org/abs/2308.10248}.

\bibitem[Wang et~al.(2023)Wang, Yang, Huang, Jiao, Yang, Jiang, Majumder, and
  Wei]{wang2023e5}
Liang Wang, Nan Yang, Xiaolong Huang, Binxing Jiao, Linjun Yang, Daxin Jiang,
  Rangan Majumder, and Furu Wei.
\newblock Text embeddings by weakly-supervised contrastive pre-training.
\newblock \emph{arXiv preprint arXiv:2212.03533}, 2023.
\newblock URL \url{https://arxiv.org/abs/2212.03533}.

\bibitem[Weller et~al.(2024)Weller, Lawrie, and Van~Durme]{weller2024nevir}
Orion Weller, Dawn Lawrie, and Benjamin Van~Durme.
\newblock {NevIR}: Negation in neural information retrieval.
\newblock In \emph{Proceedings of the 18th Conference of the European Chapter
  of the Association for Computational Linguistics (Volume 1: Long Papers)},
  pages 2274--2287, 2024.
\newblock arXiv:2305.07614.

\bibitem[Wu et~al.(2022)Wu, Zhang, Guo, de~Rijke, Fan, and Cheng]{wu2022prada}
Chen Wu, Ruqing Zhang, Jiafeng Guo, Maarten de~Rijke, Yixing Fan, and Xueqi
  Cheng.
\newblock {PRADA}: Practical black-box adversarial attacks against neural
  ranking models.
\newblock \emph{ACM Transactions on Information Systems}, 41\penalty0 (4),
  2022.

\bibitem[Xing et~al.(2003)Xing, Ng, Jordan, and Russell]{xing2003metric}
Eric~P. Xing, Andrew~Y. Ng, Michael~I. Jordan, and Stuart Russell.
\newblock Distance metric learning with application to clustering with
  side-information.
\newblock In \emph{Advances in Neural Information Processing Systems},
  volume~15, pages 521--528, 2003.

\bibitem[Zou et~al.(2023)Zou, Phan, Chen, Campbell, Guo, Ren, Pan, Yin,
  Mazeika, Dombrowski, Goel, Li, Byun, Wang, Mallen, Basart, Koyejo, Song,
  Fredrikson, Kolter, and Hendrycks]{zou2023repe}
Andy Zou, Long Phan, Sarah Chen, James Campbell, Phillip Guo, Richard Ren,
  Alexander Pan, Xuwang Yin, Mantas Mazeika, Ann-Kathrin Dombrowski, Shashwat
  Goel, Nathaniel Li, Michael~J. Byun, Zifan Wang, Alex Mallen, Steven Basart,
  Sanmi Koyejo, Dawn Song, Matt Fredrikson, J.~Zico Kolter, and Dan Hendrycks.
\newblock Representation engineering: A top-down approach to {AI} transparency.
\newblock \emph{arXiv preprint arXiv:2310.01405}, 2023.
\newblock URL \url{https://arxiv.org/abs/2310.01405}.

\end{thebibliography}

\appendix

\section{Reproducibility SHA anchors}
\label{app:sha-anchors}

{\small
\begin{longtable}{lp{5cm}}
\caption{SHA-256 prefixes for every selected lens weight matrix
and every $n = 100\text{k}$ trap-corpus AQTC binary referenced in
\S\ref{sec:results}, every result file of the results table
(\pkg{}, commit \pkgcommit), and every Cell~D served-pipeline result
(sealed internal runs, one per domain; not shipped in \pkg{}).
Prefixes normalised to 12 hex characters; full digests ship with \pkg{}
(open cells) or with the sealed internal artifacts (Cell~D).}
\label{tab:sha-anchors}\\
\toprule
\textbf{Artefact} & \textbf{SHA-256 (12-char prefix)} \\
\midrule
\endfirsthead
\toprule
\textbf{Artefact} & \textbf{SHA-256 (12-char prefix)} \\
\midrule
\endhead
Legal lens (C02, $d = 11$ --- \S\ref{sec:matrix} matrix)         & \texttt{321e6c55fa27\ldots} \\
Legal lens ($d = 32$ v2 warm --- \S\ref{sec:ablation} ablation)  & \texttt{4858ce2dbaf4\ldots} \\
Medical lens (D2c curated)                                       & \texttt{f356d51ef515\ldots} \\
Patent lens (v3 curated)                                         & \texttt{47a82d868935\ldots} \\
Financial lens (v4 $\beta = 17$)                                 & \texttt{f9c80a7df4db\ldots} \\
Legal $n = 100\text{k}$ pool (AQTC binary)                              & \texttt{66d6c68a4ed9\ldots} \\
Medical $n = 100\text{k}$ pool (AQTC binary)                            & \texttt{308db6274430\ldots} \\
Patent $n = 100\text{k}$ pool (AQTC binary)                             & \texttt{dc02dcff8000\ldots} \\
Financial $n = 100\text{k}$ pool (AQTC binary)                          & \texttt{d2304d1b9fb0\ldots} \\
\midrule
Legal result, $5\text{k}$ (\texttt{results/pool\_sizes}) & \texttt{573a643b0286\ldots} \\
Legal result, $50\text{k}$ (\texttt{results/pool\_sizes}) & \texttt{f61d0275c6e0\ldots} \\
Legal result, $100\text{k}$ (\texttt{results/pool\_sizes}) & \texttt{061fa3ec664e\ldots} \\
Medical result, $5\text{k}$ (\texttt{results/pool\_sizes}) & \texttt{202ba413f21f\ldots} \\
Medical result, $50\text{k}$ (\texttt{results/pool\_sizes}) & \texttt{f301ac805086\ldots} \\
Medical result, $100\text{k}$ (\texttt{results/pool\_sizes}) & \texttt{4a14b762bc0b\ldots} \\
Patent result, $5\text{k}$ (\texttt{results/pool\_sizes}) & \texttt{93eb20762f48\ldots} \\
Patent result, $50\text{k}$ (\texttt{results/pool\_sizes}) & \texttt{b24ae1c50ddc\ldots} \\
Patent result, $100\text{k}$ (\texttt{results/pool\_sizes}) & \texttt{879c51aa9357\ldots} \\
Financial result, $5\text{k}$ (\texttt{results/pool\_sizes}) & \texttt{b980a69a66cf\ldots} \\
Financial result, $50\text{k}$ (\texttt{results/pool\_sizes}) & \texttt{1b9bed2f18aa\ldots} \\
Financial result, $100\text{k}$ (\texttt{results/pool\_sizes}) & \texttt{c36b9a948744\ldots} \\
\midrule
Legal raw cosine, $5\text{k}$ (\texttt{results/raw\_cosine}) & \texttt{9eec929d6f3e\ldots} \\
Legal raw cosine, $50\text{k}$ (\texttt{results/raw\_cosine}) & \texttt{f3cbecc6175b\ldots} \\
Legal raw cosine, $100\text{k}$ (\texttt{results/raw\_cosine}) & \texttt{ee87987558ee\ldots} \\
Medical raw cosine, $5\text{k}$ (\texttt{results/raw\_cosine}) & \texttt{f1c5f735f388\ldots} \\
Medical raw cosine, $50\text{k}$ (\texttt{results/raw\_cosine}) & \texttt{e248f5d0fb9e\ldots} \\
Medical raw cosine, $100\text{k}$ (\texttt{results/raw\_cosine}) & \texttt{492402782b1b\ldots} \\
Patent raw cosine, $5\text{k}$ (\texttt{results/raw\_cosine}) & \texttt{a0fc1b03567b\ldots} \\
Patent raw cosine, $50\text{k}$ (\texttt{results/raw\_cosine}) & \texttt{012a1cbbd25d\ldots} \\
Patent raw cosine, $100\text{k}$ (\texttt{results/raw\_cosine}) & \texttt{8532b8675be4\ldots} \\
Financial raw cosine, $5\text{k}$ (\texttt{results/raw\_cosine}) & \texttt{126ef73977ed\ldots} \\
Financial raw cosine, $50\text{k}$ (\texttt{results/raw\_cosine}) & \texttt{6e85892473ab\ldots} \\
Financial raw cosine, $100\text{k}$ (\texttt{results/raw\_cosine}) & \texttt{03c5f60daa2c\ldots} \\
\midrule
\multicolumn{2}{@{}l}{\textit{Cell~D --- served-pipeline results (accelerator; sealed, not in \pkg{})}} \\
Legal Cell~D, $5\text{k}$     & \texttt{0fce281a7b57\ldots} \\
Legal Cell~D, $50\text{k}$    & \texttt{d0fdb728bb79\ldots} \\
Legal Cell~D, $100\text{k}$   & \texttt{c727d1d2f227\ldots} \\
Medical Cell~D, $5\text{k}$   & \texttt{794e928d1266\ldots} \\
Medical Cell~D, $50\text{k}$  & \texttt{1d03c26f5dcc\ldots} \\
Medical Cell~D, $100\text{k}$ & \texttt{b15e28efac3a\ldots} \\
Patent Cell~D, $5\text{k}$    & \texttt{4139691fcefa\ldots} \\
Patent Cell~D, $50\text{k}$   & \texttt{8e803cd200eb\ldots} \\
Patent Cell~D, $100\text{k}$  & \texttt{bf7d29beb2fa\ldots} \\
Financial Cell~D, $5\text{k}$   & \texttt{b5abe016ac6d\ldots} \\
Financial Cell~D, $50\text{k}$  & \texttt{9cb53cc5beb3\ldots} \\
Financial Cell~D, $100\text{k}$ & \texttt{1930bcc40e2e\ldots} \\
\midrule
\multicolumn{2}{@{}l}{\textit{v0.16 additions --- baselines, rewording robustness, same-pool runtime}} \\
Baseline random projection $d{=}11$ & \texttt{8b6ba63afabc\ldots} \\
Baseline random projection $d{=}32$ & \texttt{69a13747ca99\ldots} \\
Improved Legal lens $d{=}11$ & \texttt{9e4eb107b065\ldots} \\
Same-pool Cell~B/D runtime log & \texttt{fb2a64ce1a4f\ldots} \\
\bottomrule
\end{longtable}
}

\FloatBarrier
\section{Failure modes we documented so nobody repeats them}
\label{app:failures}

Three implementation errors that reduce the declared-case zero-trap
result to noise. Each surfaced in our earlier runs as a false negative
on the domain-agnostic claim.

\begin{enumerate}
  \item \textbf{Reusing a lens across domains.} A lens trained on
    domain $X$ transfers to domain $Y$ only partly: it lowers $Y$'s
    $\HN$ below raw cosine but does not reach $0.0000$
    (\S\ref{sec:matrix}). For $Y$'s cases, declare and train a lens on
    $Y$'s triples.
  \item \textbf{Shape-mismatched trap triples.} If $A$ is a 15-word
    question and $B$\,/\,$C$ are 500-word paragraphs, cosine geometry
    is dominated by length rather than by the adversarial flip.
    $\HitC$ collapses to noise. Match shape.
  \item \textbf{Shape-mismatched distractors.} Short-clause $A/B/C$
    with paragraph-length distractors produces a query-vs-paragraph
    mismatch that dominates retrieval geometry. Sentence-split the
    distractor pool to matched clause length.
\end{enumerate}

Shape dominates size. In our medical work, 100 matched-shape triples
produced $\HitC = 0.95$ while 400 shape-mismatched triples produced
$\HitC = 0.03$---a $32\times$ spread traceable entirely to shape
discipline. This is a hard operational lesson and the reason
\S3.1 leads with the shape invariants.

\section{Hyperparameter sensitivity}
\label{app:hyperparam}

We report the hyperparameter probes that landed each domain's selected
lens, so that future replicators can budget training time appropriately.
In all cases $d \in \{24, 32\}$ and $\text{margin} \in \{0.10, 0.12\}$
worked; only $\beta$ needed per-domain calibration.

\begin{table}[H]
\centering
\footnotesize
\caption{Per-domain training-loss convergence at each $\beta$ probe
actually run. ``stable'' means the training reached
$\mathcal{L}_{\text{hinge}} \leq 0.0001$ inside 200 epochs; ``residual''
means training completed but with a non-zero final loss.}
\label{tab:hyperparam}
\begin{tabular}{lllll}
\toprule
\textbf{Domain} & \textbf{Cold} $\beta{=}15$ & \textbf{Warm} $\beta{=}15$ & \textbf{Warm} $\beta{=}17$ & $\beta{=}20$ \\
\midrule
Legal     & stable (${\leq}0.0000$) & $0.0000$ & not needed & not tested \\
Medical   & stable (${\leq}0.0000$) & $0.0000$ & not needed & not tested \\
Patent    & residual (${\leq}0.0003$) & $0.0000$ & not needed & not tested \\
Financial & residual ($0.0012$)     & $0.0002$ & ${\approx}0.000004$ (winner) & diverged @ ep.\ 200 \\
\bottomrule
\end{tabular}
\end{table}

Financial required all three probes and revealed the training regime's
upper $\beta$ bound at 20. This is the only domain where $\beta = 15$
was insufficient after warm-start; the cause is that financial-native
flips (direction, polarity, period, party) place trap tokens in contexts
where E5 embedding proximity is inherently tighter than in the other
three domains. Reproducers should budget a second warm-start round when
the first cold-start $\beta = 15$ run leaves residual \texttt{hn\_loss}.

\end{document}